\RequirePackage{fix-cm}
\documentclass[twocolumn]{svjour3}          
\smartqed  

\usepackage{graphicx}
\usepackage{url}
\usepackage{xcolor}
\usepackage{hyperref}
\hypersetup{
     colorlinks=true,
     linkcolor=orange,
     filecolor=orange,
     citecolor=orange,      
     urlcolor=orange,
     }
\usepackage{color, colortbl}
\usepackage{amsfonts}
\usepackage{amsmath}
\usepackage{appendix}
\usepackage{algorithm, setspace}
\usepackage{algpseudocode}
\usepackage{csquotes}
\usepackage{amsfonts}
\usepackage{booktabs}
\usepackage{siunitx}
\usepackage{breqn}
\usepackage[normalem]{ulem}
\usepackage{tikz}
\usetikzlibrary{tikzmark}
\usepackage{multirow}

\newcommand{\p}[1]{\medskip \noindent \textbf{{#1}.}}
\newcommand{\eq}[1]{Equation~(\ref{eq:#1})}
\newcommand{\fig}[1]{Figure~\ref{fig:#1}}

\begin{document}

\title{Combining and Decoupling Rigid and Soft Grippers to Enhance Robotic Manipulation}

\author{Maya Keely$^{\dagger,1}$ \and Yeunhee Kim$^{\dagger,2}$ \and Shaunak A. Mehta$^1$ \and Joshua Hoegerman$^1$, Robert Ramirez Sanchez$^1$ \and Emily Paul$^1$ \and Camryn Mills$^2$ \and Dylan P. Losey$^{*,1}$ \and Michael D. Bartlett$^{*,2}$}

\authorrunning{Keely \textit{et al.}}

\institute{$^{1}$Collaborative Robotics Lab\\ Virginia Tech, Blacksburg, VA 24061, USA
\smallskip
\\
$^{2}$Soft Materials and Structures Lab\\ Virginia Tech, Blacksburg, VA 24061, USA
\smallskip
\\
$^{\dagger}$These authors contributed equally to this work
\smallskip
\\
$^{*}$email: \texttt{losey@vt.edu}, \texttt{mbartlett@vt.edu}
}

\maketitle

\begin{abstract}

For robot arms to perform everyday tasks in unstructured environments, these robots must be able to manipulate a diverse range of objects.
Today's robots often grasp objects with either soft grippers or rigid end-effectors.
However, purely rigid or purely soft grippers have fundamental limitations: soft grippers struggle with irregular, heavy objects, while rigid grippers often cannot grasp small, numerous items.
In this paper we therefore introduce \textbf{RISO}s, a mechanics and controls approach for unifying traditional \textbf{RI}gid end-effectors with a novel class of \textbf{SO}ft adhesives.
When grasping an object, RISOs can use either the rigid end-effector (pinching the item between non-deformable fingers) and/or the soft materials (attaching and releasing items with switchable adhesives).
This enhances manipulation capabilities by \textit{combining and decoupling} rigid and soft mechanisms. 
With RISOs robots can perform grasps along a spectrum from fully rigid, to fully soft, to rigid-soft, enabling real time object manipulation across a $1$ million times range in weight (from $2$ mg to $2$ kg).
To develop RISOs we first model and characterize the soft switchable adhesives.
We then mount sheets of these soft adhesives on the surfaces of rigid end-effectors, and develop control strategies that make it easier for robot arms and human operators to utilize RISOs.
The resulting RISO grippers were able to pick-up, carry, and release a larger set of objects than existing grippers, and participants also preferred using RISO.
Overall, our experimental and user study results suggest that RISOs provide an exceptional gripper range in both capacity and object diversity.
See videos of our user studies here: \url{https://youtu.be/du085R0gPFI}

\end{abstract}


\begin{figure*}[t!]
	\centering
	\includegraphics[width=1\textwidth]{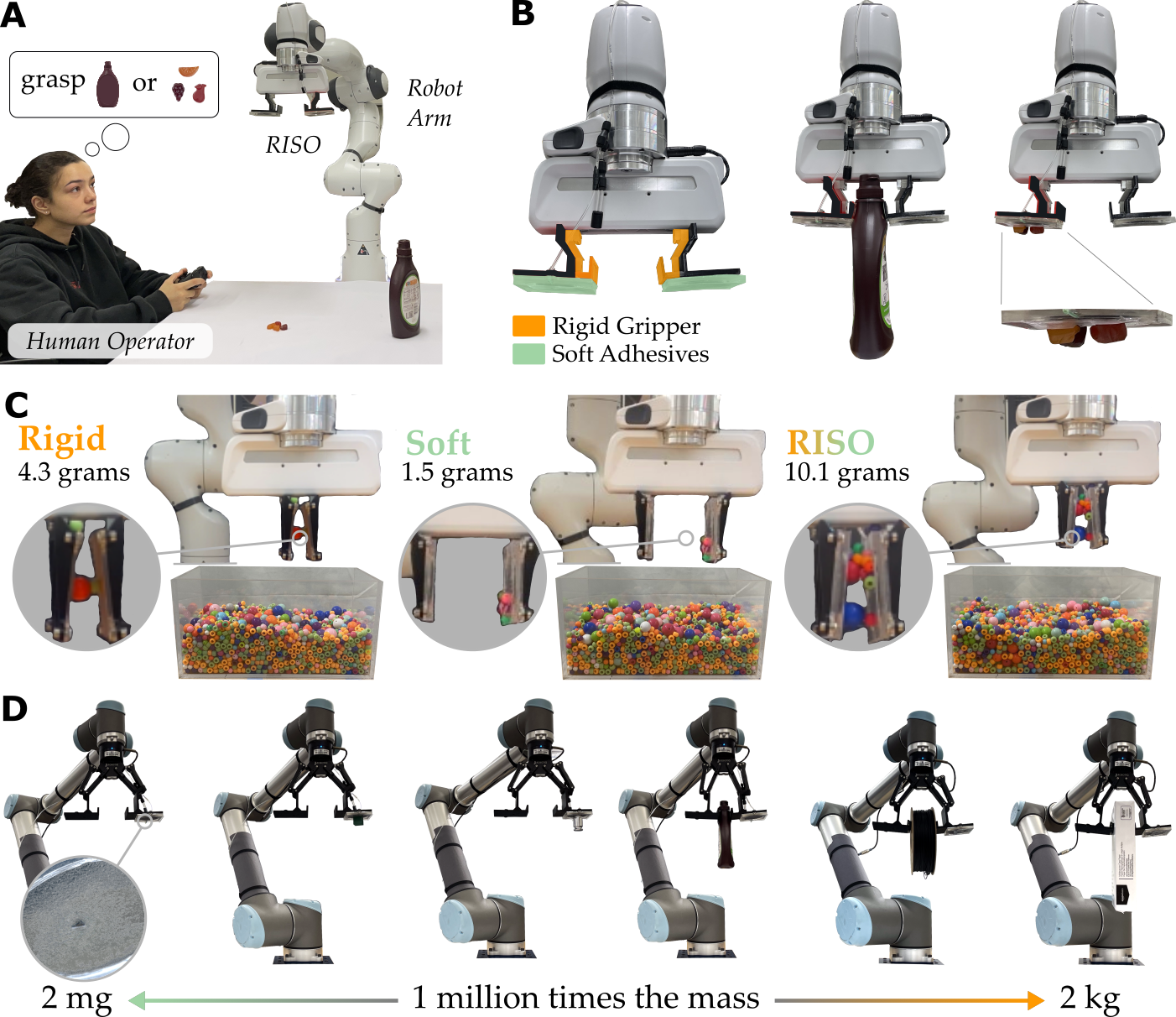}
	\caption{\textbf{RISO enhances grasping by combining and decoupling rigid and soft mechanisms.} (\textbf{A}) Human operators and robot arms can leverage RISOs (RIgid-SOft robotic grippers) to pick up, hold, and release objects. (\textbf{B}) RISOs are formed by mounting soft adhesive sheets to the surfaces of traditional rigid end-effectors. (\textbf{C}) When grasping an item RISO can use a fully rigid grasp (pinching the object between non-deformable fingers) a fully soft grasp (causing the object to adhere to its surface), or a combined rigid and soft grasp. (\textbf{D}) With this spectrum of grasps, RISO is able to pick up objects ranging from $2$ mg items to $2$ kg, a $1$ million times change in mass.}
	\label{fig:front}
\end{figure*}

\section{Introduction}

Across food processing, parts manufacturing, and assistive caregiving, there is a need for robot arms that can grasp and manipulate a diverse range of everyday objects \cite{sanneman2021state,argall2018autonomy,shintake2018soft,zhang2020state}. 
Industry-standard arms often use \textit{rigid} grippers that pinch items between two or more non-deformable fingers \cite{piazza2019century,birglen2018statistical,robotiqProductsGrippers,rojas2016gr2,webb2024wearable}.
However, these rigid grippers are not able to pick up and hold objects that are small, numerous, or irregularly shaped (e.g., a pile of candy).
To address these shortcomings, recent works have developed a variety of \textit{soft} grippers using mechanisms such as gecko-inspired adhesives \cite{ruotolo2021grasping,glick2018soft,song2014geckogripper,tian2019gecko,hao2020multimodal}, granular jamming \cite{brown2010universal,wei2016novel}, electroadhesion \cite{cacucciolo2022peeling,shintake2016versatile,alizadehyazdi2020electrostatic}, or responsive materials \cite{tatari2018dynamically,barron2023unified,wan2023tunable,song2014soft,song2017controllable,coulson2022versatile,hu2021soft}.
But soft grippers have a separate grasping domain, and are often unable to hold the large, heavy items that rigid end-effectors are able to pick up and carry (e.g., a bottle of syrup).
When a robot arm leverages grippers that are either purely rigid or purely soft, it fundamentally limits the types of objects that robot can manipulate.

We therefore seek to expand the range of graspable objects by combining rigid and compliant components within the robot's gripper.
Recent works have started to develop grippers that integrate both elements \cite{chen2023soft,hernandez2023current,nasab2017soft,park2018hybrid,li2022stiffness,peng2024enhanced,gafer2020quad,keely2024kiri,guo2020self}. 
However, within existing designs the rigid and soft capabilities are inherently \textit{coupled}; e.g., robotic fingers that alternate between rigid links and soft joints.
By contrast, our core insight is that we can \textit{decouple} and \textit{couple} the rigid and soft mechanisms by combining traditional rigid end-effectors with soft adhesive sheets.
This combination and decoupling has the potential to fundamentally increase the range of robotic grasps: from purely rigid (e.g., manipulating a bottle of syrup) to purely soft (e.g., picking up a pile of candy) to a combination of rigid and soft (e.g., holding beads of multiple sizes).

We leverage our insight to create \textbf{RISOs}, robotic grippers that unify \textbf{RI}gid end-effectors with \textbf{SO}ft materials (see \fig{front}A-B).
RISOs are formed by mounting sheets of switchable adhesives onto the surfaces of rigid robotic fingers, and then controlling both the rigid and soft components during manipulation tasks.
When the robot arm moves to pick up an item, it can (a) pinch the object between the fingers of the rigid gripper, (b) cause the object to stick to the soft surfaces, or (c) use a combination of the rigid pinch and soft adhesion to hold the object (see \fig{front}C).
In practice, RISO advances the range of robotic grasping by combining and decoupling purely rigid and purely soft grippers.
For example, RISOs can grasp objects in real-time across a \textit{one million times range in weight} --- picking up, carrying, and releasing objects from $2$ mg to $2$ kg (see \fig{front}D).

We first explored the feasibility of RISO in preliminary work \cite{mehta2023riso}.
In this paper, we now make the following contributions towards a unified framework for rigid-soft RISO grippers:

\p{Characterizing gripper capabilities}
Functionally, our soft materials take the form of flat silicone sheets that we can rapidly control to switch between grasping and releasing in less than $0.1$ seconds.
We characterize the gripping force applied by these structures as a function of the grasped object's radius, curvature, roughness, and porosity.
We find that the soft materials combine both adhesion (i.e., using surface forces to attach to the object) and wrapping (i.e., constricting around the object), achieving forces reaching $50$~N in idealized testing conditions.

\p{Integrating RISOs with robot arms and human users}
We recognize that increased gripping capability is not meaningful if robot arms and human operators cannot harness RISOs in everyday tasks.
Accordingly, we develop control strategies with varying levels of autonomy to make it easier for robots and humans to utilize RISOs.
In particular, we present a shared autonomy approach that (a) uses the human's joystick inputs to infer their desired object and grasp type, and then (b) autonomously aligns the RISO to complete the intended grasp.
We show that this shared autonomy method reduces the number of human inputs needed to complete manipulation tasks with RISOs.

\p{Comparing to existing grippers} 
We perform grasping experiments with robot arms and human operators to compare RISOs, an industrial soft gripper, and a granular jamming gripper. 
Across both autonomous and human controlled conditions, we observe that robots equipped with RISOs are able to pick up, hold, and release a more diverse range of everyday objects. 
Users also indicate that they subjectively prefer using RISOs over the state-of-the-art alternatives.

\p{Demonstrating practical applications} 
We finally showcase our system's practical abilities in assistive scenarios by controlling a robot arm and RISO gripper to make pizza. 
Here the human teleoperates the arm and RISO to manipulate larger food objects (e.g., placing the crust, spreading the sauce) and arrange smaller food items (e.g., pepperoni, pepper, and olive toppings).
When viewed together, our results suggest that RISOs provide an exceptional gripping range in both capacity and object diversity, clearly distinguishing our framework from the baselines. These advances provide a path forward for robotic grippers, particularly within unstructured environments where manipulation of diverse and unexpected items is essential.

\section{Materials and Methods} \label{sec:methods}

Here we present our mechanics and control approach for RISO grippers.
In Section~\ref{sec:mechanics} we explain the physics behind the soft adhesives, as well as their fabrication process.
We then form RISOs by mounting sheets of these pneumatically-actuated soft materials onto the surfaces of rigid end-effectors.
This combination of a rigid gripper and adhesives sheets introduces new control variables: e.g., deciding whether to use the soft adhesives or rigid grippers to pick up an object.
In Section~\ref{sec:control} we partially automate these decisions and make it easier for humans and robot to control RISOs by developing a share autonomy approach for RISO grippers.

\subsection{Creating RISO Grippers} \label{sec:mechanics}

To understand the adhesive capacity of the soft gripper, we start with the underlying physical relationship between adhesive force $F_c$, contact area, and gripper compliance \cite{bartlett2012looking,croll2019switchable}:
\begin{equation}\label{eq:adhscaling}
F_{c}\sim\sqrt{G_{c}}\sqrt{\frac{A}{C}}
\end{equation}
Here $A$ is the contact area between the soft gripper and the target object, $C$ is the gripper's compliance in the loading direction, and $G_c$ captures the fracture energy of the interface. In practice, $G_c$ can treated as the energy per unit area that needs to be applied to separate the interface between gripper and object. In the case of reversible adhesives that primarily rely on van der Waals interactions \cite{autumn2002evidence,lee2023bioinspired,bartlett2023peel}, higher $G_c$ can be achieved through viscoelastic dissipation or other lossy processes at the interface.

Of particular interest for adhesive-based object manipulation is the ability to dramatically increase $F_{c}$ when gripping an object, and then rapidly decrease $F_{c}$ when releasing that object.
To maximize $F_{c}$, \eq{adhscaling} demonstrates that $A$ should be increased, $C$ should be decreased, and $G_c$ should be high. Typically, to achieve high contact area $A$ between the gripper surface and the surface of the object, the gripper should be as \textit{soft} as possible. However, to minimize compliance $C$, the gripper also needs to be as \textit{stiff} as possible. To accommodate these contrasting design directions, we focus on the overall $A/C$ ratio between contact area and compliance. 
We propose an adhesive design that is soft during initial contact between gripper and object (increasing contact area $A$), and then becomes rigid while maintaining that same contact area (decreasing the compliance $C$).
Accordingly, to achieve a large $A/C$ ratio, it is desirable for the adhesive to be \textit{tunable} so that the compliance can be changed during contact to first increase $A$ and then decrease $C$. 

We apply these adhesion principles to create the soft mechanism for RISO grippers. This adhesive consists of a tunable elastomeric membrane on a foam foundation (see \fig{SMAdhFab} for fabrication details). We actively control the membrane through pneumatic pressure (see \fig{SMAdhSetup}), allowing for three primary states: neutral pressure where the membrane is flat, positive pressure when the membrane is inflated (increasing compliance), and negative pressure where the membrane is constricted (decreased compliance). Grasping and releasing items is achieved by switching between these states.

\p{Release} To release an item we want to minimize $F_c$. Our soft adhesive achieves this by switching from negative pressure to positive pressure and inflating the membrane (see \fig{SMAdhRelease}). This inflation causes the contact area to decrease, minimizing the ratio $A/C$.

\p{Neutral to Negative} To increase $F_c$ and adhere to a target object, our soft gripper can use two different processes. The first approach is \textit{neutral to negative}, where the membrane is initially at atmospheric pressure. We start the grasp by bringing the membrane into contact with the object, and then apply a negative pneumatic pressure to increase the stiffness of the membrane and foundation. This decreases compliance $C$ while having a negligible effect on contact area $A$, increasing overall gripper force $F_{c}$ (see \fig{indentertest}A).

\p{Positive to Negative} To further increase the grasping force, we can alternatively start the membrane at a positive pressure (i.e., the membrane is initially inflated). Here the membrane is first inflated into a hemispherical shape and then pressed into the object to be grasped. This causes the object to embed into the inflated membrane and maximizes $A$ by creating both side and normal contacts. The pressure inside the membrane is then decreased, causing the membrane to tightly constrict around the object. This minimizes $C$ during the loading phase while also activating other debonding mechanisms --- like frictional sliding --- that can increase $G_c$ (see \fig{indentertest}B).

To create RISOs, one or more of these soft adhesive sheets are mounted onto the surfaces of rigid robotic grippers. Because the soft adhesives only change shape on their surface and not throughout their volume, they can be readily incorporated onto end-effectors without loss of functionality. 
Similarly, the rigid grippers can actuate normally as they are not encumbered by soft adhesive sheets. 
This mechanical combination and decoupling provides the fundamental advantage of our design: we can have purely rigid, purely soft, or combined rigid/soft grasps depending on where the soft adhesives are mounted and how the overall RISO is controlled.

\subsection{Controlling RISO Grippers} \label{sec:control}

RISO has the potential to enhance grasping capabilities by combining and decoupling rigid and soft mechanisms.
But to harness these capabilities, we must integrate RISOs with robot arms and human operators.
The key challenge here is that RISOs introduce new low-level variables (e.g., the pressure of the soft adhesives) and high-level decisions (e.g., choosing to grasp an object with the soft materials, the rigid gripper, or both). 
Below we formulate this controls problem for RISO grippers.
We then introduce solutions with varying levels of automation: from fully automated, to fully human controlled, to a shared autonomy approach between the human and robot.

\p{Formulating the Control Problem}
We consider settings where a RISO gripper is the end-effector of a robot arm.
Let $s \in \mathcal{S}$ be the system state, where $s$ includes the joint position of the robot arm $s_{robot}$, the pose of the rigid gripper $s_{\text{rigid}}$, and the pose of $n$ soft adhesives $s_{\text{soft}_1}, \ldots, s_{\text{soft}_n}$.
To pick up and manipulate objects, the robot arm must move the RISO so that either the rigid gripper $s_{\text{rigid}}$ or one of the soft adhesives $s_{\text{soft}_i}$ is in contact with the desired item.
We use $o \in \mathcal{O}$ to denote the position of objects in the environment; our robot arm observes each object $o$ by using a mounted RGB-D camera and the Yolo-v5 object detection algorithm \cite{jiang2022review}.

There are two inputs used to move the robot arm and actuate the RISO gripper.
First, the robot can take autonomous actions $a_{\mathcal{R}} \in \mathcal{A}$ to change its own joint velocity.
Second --- if a human operator is present --- this human can teleoperate the robot arm and RISO gripper with a joystick (see Figure~\ref{fig:front}A).
The direction the user presses on the joystick sends a corresponding joint velocity command $a_{\mathcal{H}} \in \mathcal{A}$ to the robot.
The objective for the system is to reach for and manipulate objects. 
Let $o^* \in \mathcal{O}$ be the target object, and let $g^* \in \mathcal{G}$ be the target grasp type.
For example, if the human wants to pick up small, numerous candies, then $g^*$ could be the soft adhesive; alternatively, if the robot arm it trying to pick up a bottle of syrup, then $g^*$ should be the rigid gripper.
Our controls problem is identifying an efficient sequence of autonomous robot actions and/or human inputs that will cause the robot to pick up object $o^* \in \mathcal{O}$ using the RISO grasp type $g^* \in \mathcal{G}$.

\p{Shared Autonomy for RISO Grippers}
We developed three control strategies for RISO grippers along a spectrum of autonomy. 
At one extreme is fully \textbf{autonomous} control: here the robot arm acts in isolation without any human guidance.
The robot leverages its RGB-D camera to detect objects in the workspace, and then autonomously moves the arm with actions $a_{\mathcal{R}}$ to align RISO with the closest object.
RISO determines whether to attempt grasps using the rigid mechanism or soft adhesives based on the perceived size of the object --- for objects with a height greater than $75$~mm RISO uses the rigid gripper.
At the other extreme of the spectrum is fully \textbf{human} control: in this condition humans use a joystick to teleoperate the robot arm and RISO gripper throughout the task.
The robot simply executes the human's commanded actions $a_{\mathcal{H}}$ to move its end-effector. 
When attempting to perform a soft grasp, human users must teleoperate the RISO to directly adjust low-level variables including the normal force and gripper pressure (i.e., how hard the gripper presses down on the target object and how much the pneumatic membrane is inflated).

Between these robot and human extremes we propose a \textbf{shared autonomy} framework for RISO grippers.
This approach is designed to integrate the human, robot, and RISO while reducing the human's workload.
Intuitively, under our approach the robot (a) infers the desired object $o^*$ and grasp $g^*$ from the human's inputs, and then (b) partially automates the process of reaching for and grasping that item.
To estimate the human's desired object and grasp given the previous states $s^{0:t}$ and human actions $a_{\mathcal{H}}^{0:t}$, the robot leverages Bayesian inference:
\begin{equation*}
    P(o, g \mid s^{0:t}, a_{\mathcal{H}}^{0:t}) = b^{t+1}(o, g) \propto P(a_{\mathcal{H}}^t \mid s^t, o, g) \cdot b^t(o, g)
\end{equation*}
where $P(a_{\mathcal{H}}^t \mid s^t, o, g)$ models the human as a noisily-rational agent \cite{jeon2020reward}, and belief $b(o, g)$ is the joint probability that the human wants to reach object $o$ with grasp $g$.
For instance, if the human teleoperates the robot to move RISO's soft adhesive towards a pile of candies, then $b(o, g)$ should report that the candies are the most likely object and the soft adhesives are the most likely grasp.
The robot chooses actions $a_{\mathcal{R}}$ to assist the human in real time based on this inferred belief:
\begin{equation} \label{eq:ar}
    a_{\mathcal{R}} = \sum_{o \in \mathcal{O}} \sum_{g \in \mathcal{G}} (o - s_g) \cdot b(o, g)
\end{equation}
In practice, \eq{ar} causes the robot arm to autonomously move towards likely objects while aligning the RISO gripper for the human's preferred grasp.
This leads to a shared autonomy framework where the human is responsible for high-level decisions (e.g., which object to grasp) and the robot assists with the low-level variables (e.g., moving the robot arm directly above the object, and automating the pressure changes for the soft adhesive).
Our resulting shared autonomy approach is distinct from related works such as \cite{jain2019probabilistic,javdani2018shared,losey2022learning,jonnavittula2022sari} because we not only need to infer the human's goal object, but also how the human wants to use the RISO to grasp that object.
Our approach leads to a shared autonomy framework that coordinates both the robot arm and RISO gripper to complete manipulation tasks.

\section{Results} \label{sec:results}

In this section we conduct experiments and user studies to explore the capabilities of RISO grippers.
We start by exploring how adhesion and wrapping mechanisms contribute to the soft material's overall gripping force (Section~\ref{sec:r1}), and then characterize gripping forces as a function of the target object's radius, curvature, roughness, and porosity (Section~\ref{sec:r2}).
Equipped with this understanding of the soft components, we next apply RISOs in systems with robot arms and human operators.
In Section~\ref{sec:r3} we compare the range of objects that robots can grasp when using RISO grippers, industrial soft grippers, and a granular jamming gripper.
We also conduct user studies to assess how easy it is for robots and humans to utilize RISOs with autonomous control, human control, and shared autonomy (Section~\ref{sec:r4}).
Put together, these four experiments characterize the soft components of RISOs, compare RISOs to state-of-the-art alternatives, and demonstrate how our controllers improve gripper performance and ease-of-use.

\begin{figure*}[ht!]
	\centering
	\includegraphics[width=\textwidth]{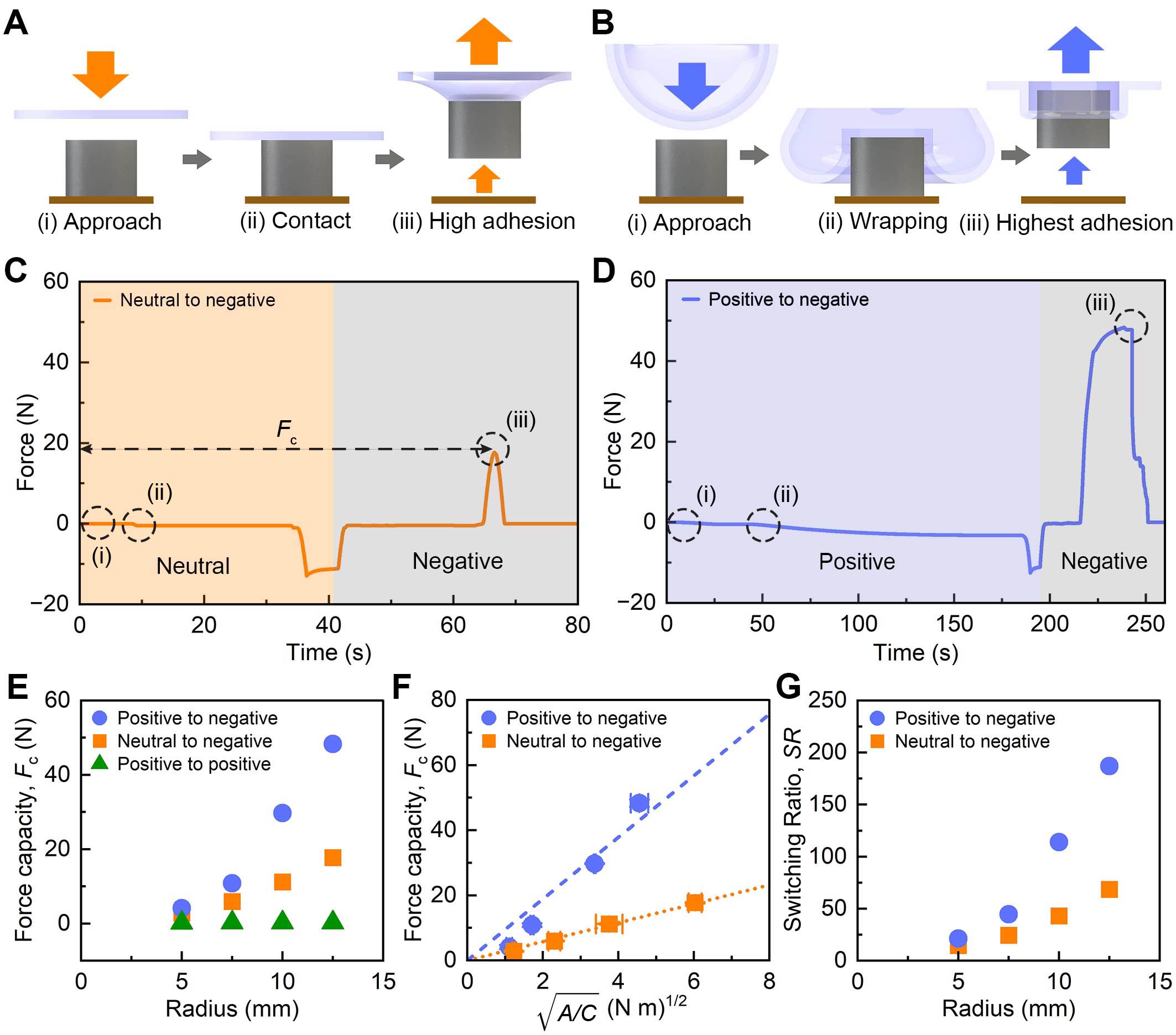}
	\caption{\textbf{Measuring the force capacity of RISO's soft adhesives.} 
    The soft adhesives operate under the principles from \eq{adhscaling}, and seek to maximize contact area while minimizing surface compliance.
    (\textbf{A}) Soft grasps while switching the membrane from neutral to negative pressure. 
    (\textbf{B}) Soft grasps while switching the membrane from positive to negative pressure. 
    (\textbf{C}) Force profiles for neutral to negative and (\textbf{D}) positive to negative with a $12.5$ mm smooth indenter (circles represent testing stages from \textbf{A} and \textbf{B}.) 
    (\textbf{E}) Force capacity $F_{c}$ vs. indenter radius.
    (\textbf{F}) Force capacity $F_{c}$ vs. $\sqrt{A/C}$, where the points represent the experimental data and the lines represent the prediction from \eq{adhscaling}. Here $G_c$= 4.2 J/m${}^{2}$ and $G_c$= 44.7 J/m${}^{2}$ for the slopes of the lower and upper lines.
    (\textbf{G}) Adhesion switching ratio ($SR$) as a function of indenter radius.   
	}
	\label{fig:indentertest}
\end{figure*}

\subsection{Measuring the Force Capacity of RISO's Soft Adhesives}\label{sec:r1}

We start by measuring the soft adhesive's force capacity when grasping a indenter (e.g., a target object) under controlled conditions.
Figures~\ref{fig:indentertest}C and \ref{fig:indentertest}D show the force profiles for a $12.5$ mm smooth indenter. 
To conduct these tests, the soft membrane was first brought into contact with the indenter, and then the membrane state was switched to a negative pressure. We finally pulled the soft gripper and indenter apart from one another until the two separated, enabling us to measure the total adhesive force $F_c$ (see \fig{SMAdhTest}).
Under the \textbf{positive to negative} condition the membrane was initially inflated (Figure~\ref{fig:indentertest}B), while for the \textbf{neutral to negative} condition the membrane started at atmospheric pressure (Figure~\ref{fig:indentertest}A).

Figure~\ref{fig:indentertest}E extends our results across indenters of different radii.
To provide a baseline for these results, we included a \textbf{positive to positive} condition where the membrane was always inflated.
As expected, in this positive to positive case the adhesion was very low for all indenter radii; because the membrane never decreased its compliance, it was unable to achieve higher $F_c$ values.
Next, we increased the adhesion force by switching between neutral and negative states. Here we observed higher $F_c$ as the indenter radii increased (i.e., as the contact area increased), and the soft gripper was able to reach a grasping force of $18$ N for the $12.5$ mm indenter.
We finally repeated these same tests with the positive to negative condition. As before, the force capacity $F_c$ increased with the radius of the target object, but now the grasping forces were larger because of the combination of adhesive and wrapping mechanisms (Figure~\ref{fig:indentertest}B).
In the positive to negative condition the soft adhesive reached a maximum gripping force of $50$ N for an indenter radius of 12.5 mm.

Based on the adhesive scaling analysis from \eq{adhscaling}, we plot $F_{c}$ vs. $\sqrt{A/C}$ in \fig{indentertest}F and find a good agreement between the physical model and experimental data. 
For the neutral to negative condition the slope gives a fracture energy of $G_c$= $4.2$ J/m${}^{2}$, which is in line with previous results \cite{swift2020active}.
By contrast, the positive to negative case produced a significantly larger fracture energy of $G_c$= $44.7$ J/m${}^{2}$. 
This suggests that the wrapping mechanism enhanced the capabilities of the soft adhesive by increasing $G_c$.
We attribute this effect to the frictional sliding and hydrostatic pressure that the membrane applies to the target object when it is wrapped around that object.

Another important parameter for manipulation is the adhesion switching ratio, $SR \equiv F_{high}/F_{low}$, where $F_{high}$ is the adhesive force in the ``on'' state (i.e. neutral to negative or positive to negative) and $F_{low}$ is the adhesive force in the ``off'' state (i.e. positive to positive).
We calculate this for our system where the $F_{high}$ is the adhesive force for the adhesion and wrapping modes, while the $F_{low}$ is the adhesive force in the release condition. 
We find that the adhesion switching ratio increases with increasing indenter radii (\fig{indentertest}G), reaching a maximum value of $187$ for the $12.5$ mm indenter. Overall, the soft adhesive's ability to achieve a high force capacity ($\sim$~$50$~N), rapid grasping and release ($<0.1$ s), and high switching ratio provides key properties for object manipulation.

\begin{figure*}[ht!]
	\centering
	\includegraphics[width=1\textwidth]{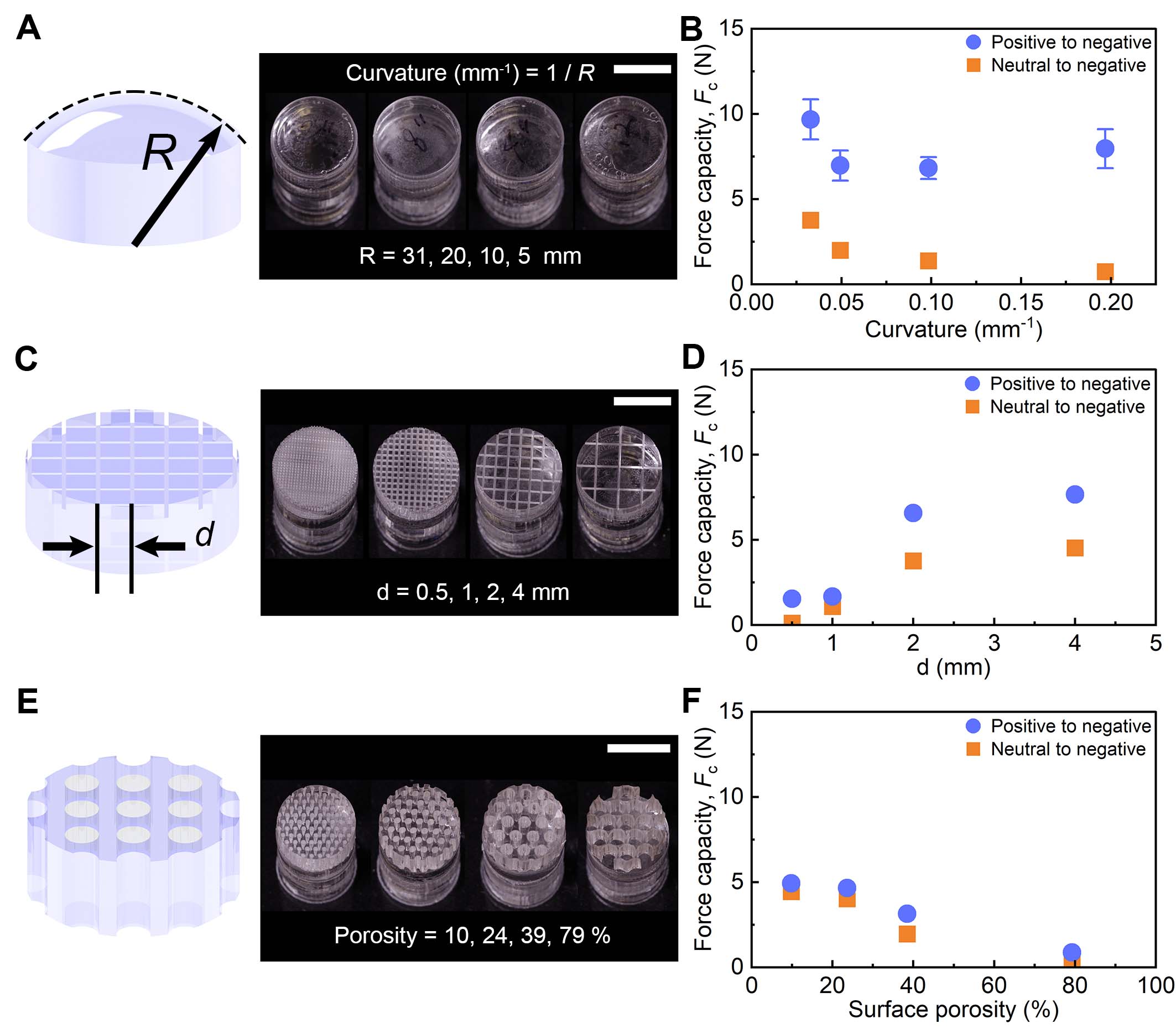}
	\caption{\textbf{Characterizing RISO's soft adhesives across diverse objects.} 
 (\textbf{A}) Curved indenters with four different curvatures. 
 (\textbf{B}) Force capacity $F_{c}$ vs. indenter curvature. 
 (\textbf{C}) Rough indenters with lines etched at distance $d$ from one another. 
 (\textbf{D}) Force capacity $F_{c}$ vs. different line distances. 
 (\textbf{E}) Porous indenters with four levels of porosity. 
 (\textbf{F}) Force capacity $F_{c}$ vs surface porosity. Across all plots the indenters have a radius of $7.5$ mm and the scale bars are $10$ mm.
	}
	\label{fig:differentsurface}
\end{figure*}

\subsection{Characterizing RISO's Soft Adhesives across Diverse Objects} \label{sec:r2}

In Section~\ref{sec:r1} we focused on idealized objects to explore the mechanics behind the soft adhesives. We next move towards diverse objects, and characterize how RISO's soft adhesive can grasp items as a function of their curvature, roughness, and surface porosity (see \fig{differentsurface}). 
For each type of object we measured the gripper's force capacity when using the neutral to negative grasp (i.e., pure adhesion) as well as the positive to negative grasp (i.e., combined adhesion and wrapping).

We start with curved indenters in \fig{differentsurface}A-B. Here the soft adhesive's performance depended on the actuation strategy: in the neutral to negative case $F_c$ decreased as the curvature increased, while with the positive to negative pressure change $F_c$ was roughly constant. Intuitively, we might expect the force capacity to be lower for more curved items because of a decreased surface area (i.e., the soft adhesive was only making contact with the tip of the indenter). The wrapping effect within the positive to negative condition mitigated this issue and enhanced the force capacity across all curvature values.
We next tested target objects with engraved surfaces to create effective, controlled roughness (see \fig{differentsurface}C-D).
When the engraved lines were closer together the surface was more rough --- leading to lower contact area $A$ and smaller $F_c$ values.
Conversely, when the lines were far apart the surface was more smooth, resulting in larger contact area and gripper forces. Here we observed a small improvement of the positive to negative over the neutral to negative condition.
Finally, we measured the adhesive force when grasping objects of different surface porosity (see \fig{differentsurface}E-F).
As the porosity increased the surface area decreased --- this led to lower force capacity across both gripper conditions.
Nonetheless, even with a surface porosity of nearly $80\%$, RISO's soft gripper was able to generate adhesion forces of up to $0.9$ N. 
We note that our soft adhesive is not applying a vacuum on the target objects; instead, we are grasping these porous objects by adhering to their remaining surface area.

\begin{figure*}
    \centering
    \includegraphics[width=0.99\textwidth]{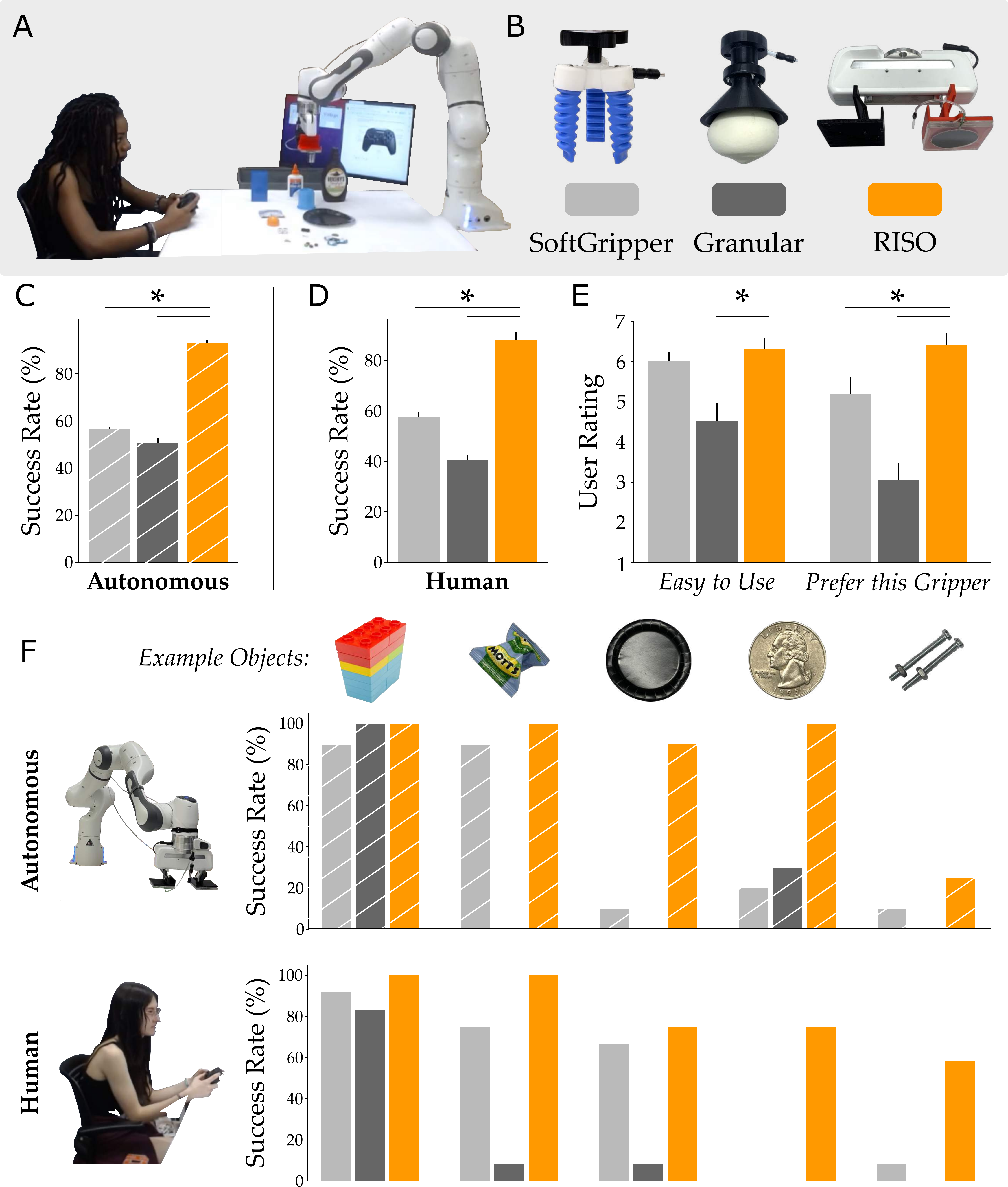}
    \caption{\textbf{Comparing RISOs to existing grippers.} (\textbf{A}) Experimental setup. Grippers were attached to a $7$-DoF robot arm and used to grasp, move, and drop a dataset of $15$ household objects. 
    (\textbf{B}) We compared an industrial SoftGripper, a granular jamming gripper (Granular), and RISO. 
    (\textbf{C}) Success rates for each gripper when the system was fully automated. 
    (\textbf{D}) Success rates for each gripper with a human-in-the-loop. A total of $12$ participants remotely controlled the robot arm and grippers using a joystick.  
    (\textbf{E}) After working with each gripper users responded to a $7$-point Likert scale survey. Users indicate how easy it was to use the gripper and which grippers they preferred.
    (\textbf{F}) Success rates for $5$ sample objects where RISO outperformed SoftGripper and Granular. For additional results on this user study see Tables~\ref{table:collab3}--\ref{table:collab2}.}
    \label{fig:user1}
\end{figure*}

\subsection{Comparing RISOs to Existing Grippers} \label{sec:r3}

Over the previous two experiments we isolated the soft component of RISO grippers under idealized conditions.
Moving forward, we will unify these soft materials with rigid end-effectors to form RISO grippers.
Our core insight is that RISOs \textit{combine and decouple} the rigid and soft gripper mechanisms, enabling grasps along a spectrum from purely rigid, to purely soft, to rigid-soft.
We hypothesize that this decoupling will fundamentally increase the range of objects that RISOs can pick up, hold, and release.
To test this hypothesis, here we explore the range of objects that robot arms and human operators can grasp with RISO grippers and state-of-the-art alternatives.

\p{Independent Variables} We compared three different grippers: RISO, a industrial SoftGripper \cite{DobotMagician}, and a granular jamming gripper \cite{brown2010universal} (see \fig{user1}B and \fig{collab1}A).
Each of these grippers were mounted at the end-effector of a $7$-DoF FrankaEmika robot arm \cite{frankaFrankaEmika}.

To understand how these different grippers performed in isolation --- without any human guidance --- we applied our autonomous controller from Section~\ref{sec:control}.
Here the robot arm identified objects with a camera, and then autonomously moved its gripper to align with the detected object.
Next, to test how the grippers performed with a human-in-the-loop, we applied the human controller from Section~\ref{sec:control}.
Under this approach human participants operated a joystick to remotely control the robot arm and attached grippers without any autonomous assistance. 
We recruited $12$ participants ($4$ female, average age $22.9 \pm 2.73$) from the Virginia Tech community to take part in this study. 
All participants provided informed written consent following university guidelines (IRB $\#22$-$308$).
The order in which participants used the grippers was randomized: some started with RISO, others started with a SoftGripper, and others started with the granular jamming gripper.

\p{Dependent Variables} The gripper, robot arm, and human operator collaborated to pick up, carry, and then drop a dataset of $15$ household objects (see \fig{user1}A and \fig{collab1}B).
During each interaction the system manipulated one item: the interaction was considered a success if the gripper picked up the target object, carried it across the table, and then dropped it in the bin.
To measure the user's perception of each gripper, we also conducted a $7$-point Likert scale survey within the human-in-the-loop condition.
Our questions focused on how easy it was to use the gripper, and whether participants preferred one gripper over the alternatives (see \fig{user1}E and Table~\ref{table:collab3}).

\p{Autonomous Results}
\fig{user1}C shows the results of the autonomous test.
This experiment compared the performance of RISO, the SoftGripper, and a granular jamming gripper when the robot's behavior was fully automated.
Across all objects and $10$ autonomous trials, the manipulation success rates were SoftGripper: $56\%$, Granular: $51\%$, and RISO: $93\%$ (see Table~\ref{table:collab2}).
These differences in performance were statistically significant.
A repeated measures ANOVA revealed that the gripper type had a significant effect on the overall success rate across $10$ trials ($F(2,18)=156.42$, $p<0.05$), and post-hoc tests in \fig{user1}C confirmed that RISO had a higher success rate than either the pneumatic SoftGripper ($p<0.05$) or the granular jamming gripper ($p<0.05$).

\p{Human-in-the-Loop Results}
\fig{user1}D summarizes our results across $12$ human participants.
This experiment compared the performance of RISO to state-of-the-art alternatives when the system was fully controlled by human operators.
We again found that the gripper type had a significant effect on the grasping success rate ($F(2, 22)=138.37$, $p<0.05$).
Participants successfully grasped and manipulated objects more often with RISO as compared to the pneumatic SoftGripper ($p<0.05$) and the granular jamming gripper ($p<0.05$).
Across $12$ users the success rates were SoftGripper: $58\%$, Granular: $41\%$, and RISO: $88\%$ (see Table~\ref{table:collab2}).
When surveyed, participants also expressed a subjective preference for the RISO gripper (see \fig{user1}E and Table~\ref{table:collab3}).
Participants perceived the RISO gripper as easier to use than the granular jamming gripper ($p<0.05$),
and indicated that they would rather use the RISO gripper again in the future instead of either the pneumatic SoftGripper ($p<0.05$) or granular jamming gripper ($p<0.05$).

\p{Comparing Grippers}
Whether the grippers were controlled directly by humans or autonomously by the robot arm, the RISO gripper was able to grasp, carry, and drop a more diverse range of items than the SoftGripper or granular jamming gripper.
In \fig{user1}F we break down these results for some sample objects in the dataset (see all object success rates in Table~\ref{table:collab2}).
Across both human and autonomous control, the pneumatic SoftGripper failed to pick up small, numerous items (e.g., the nuts) or items that slip through the soft fingers (e.g., the quarter).
Similarly, the granular jamming gripper struggled to grasp flat objects that the gripper could not wrap around (e.g., the plate).
RISO was able to overcome these limitations through its combined and decoupled design: the robot could use the soft adhesive sheets to pick up small, numerous items, and the rigid end-effector to grasp large, heavy objects.

\begin{figure*}
    \centering
    \includegraphics[width=2\columnwidth]{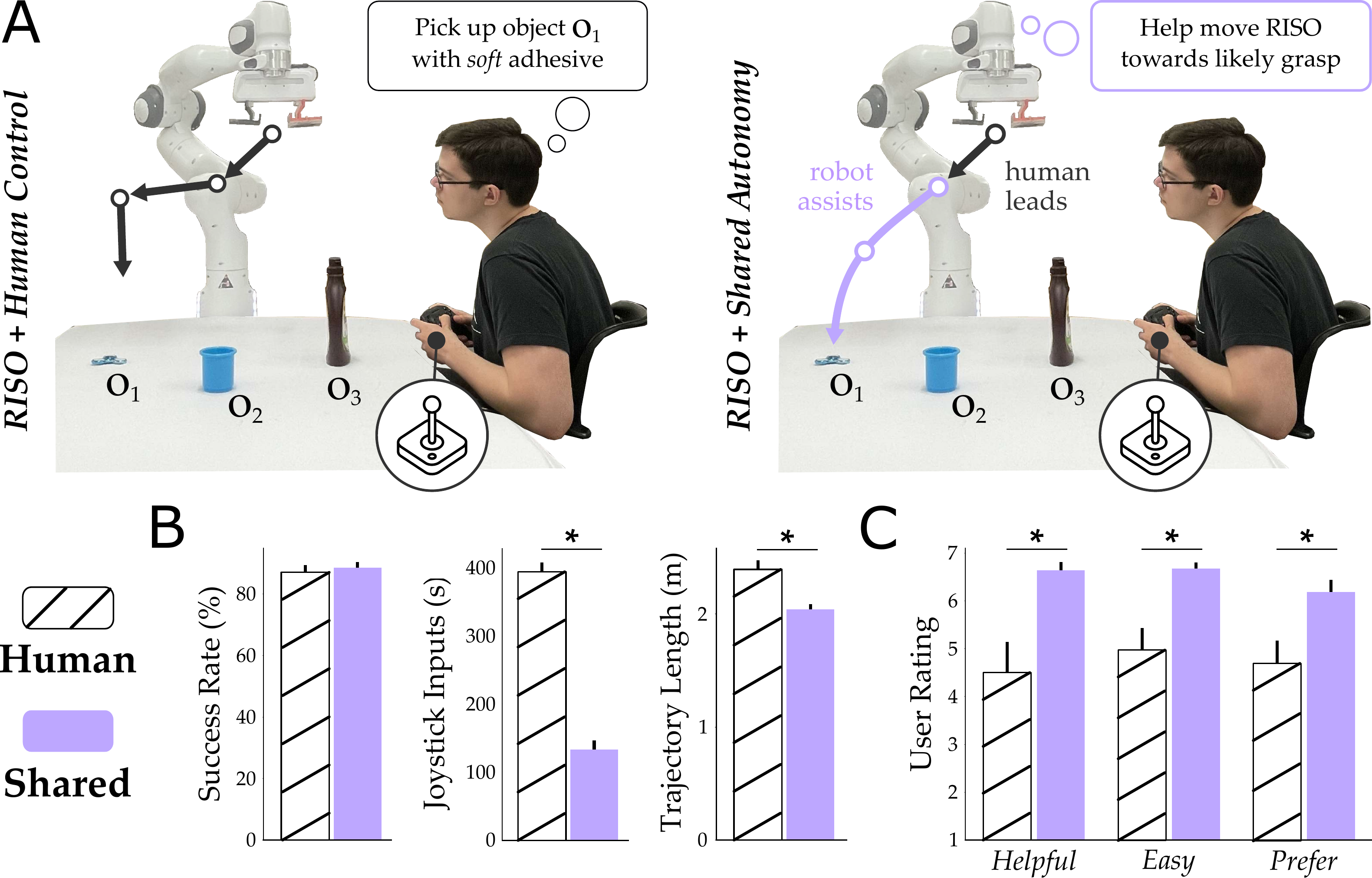}
    \caption{\textbf{Making it easier for humans to utilize RISOs.}  (\textbf{A}) Overview of human control and shared autonomy. In human control the user teleoperates the robot and RISO throughout the entire manipulation task. By contrast, in shared autonomy the system uses the human's inputs to infer their desired object and grasp type. The system then partially automates the robot arm and RISO to help complete that grasp. (\textbf{B}) Objective results from a user study with $12$ participants. With shared autonomy users were able to complete grasps with fewer joystick inputs and shorter robot trajectories. (\textbf{C}) Users responded to a $7$-point Likert scale survey to indicate how \textit{helpful} the controller was, how \textit{easy} it was to use the gripper, and which control approach they \textit{prefer} to use. Higher scores indicate agreement (e.g., more helpful), and an $*$ denotes statistical significance.}
    \label{fig:user2}
\end{figure*}

\subsection{Reducing Effort with Shared Autonomy} \label{sec:r4}

When comparing RISO to state-of-the-art grippers in Section~\ref{sec:r3} we tested the extremes of our controls spectrum: fully autonomous or fully human controlled.
Both of these control modes have advantages. 
On the one hand, an autonomous robot can optimize the low-level variables introduced by RISO grippers (e.g., the normal force and pneumatic pressure).
On the other hand, a human can specify high-level targets and grasps (e.g., selecting whether to use the rigid, soft, or rigid-soft gripping mechanism).
In Section~\ref{sec:control} we combined the benefits of both approaches under a \textit{shared autonomy} method for RISO grippers.
This method integrates the human, robot arm, and RISO gripper: based on the human's joystick inputs, the robot and RISO infer the human's desired object, and then automate the arm and gripper behavior to help complete the task.

\p{Independent Variables} We compared our proposed shared autonomy method to systems completely controlled by the human.
Twelve new participants from the Virginia Tech community ($2$ female, average age $24.1 \pm 4.5$ years) provided informed consent and took part in this user study (IRB $\#22$-$308$).
These participants interacted with a joystick to remotely control the $7$-DoF FrankaEmika robot arm \cite{frankaFrankaEmika} and its attached RISO gripper.
Within the \textit{human} condition the robot directly executed the user's commands without providing any assistance.
By contrast, under \textit{shared autonomy} the robot and RISO provided partial assistance based on the inferred object and grasp type.
The order of these conditions was counterbalanced: half of the users started with shared autonomy, and the other half started with human control.

\p{Dependent Variables} Each new participant operated the robot and RISO to pick up, carry, and then drop the same $15$ household items as in Section~\ref{sec:r3} (also see \fig{collab1}B).
Our underlying hypothesis was that shared autonomy would make it easier for humans to leverage the RISO gripper.
To measure changes in efficiency, we recorded the amount of time users actively controlled the robot with joystick inputs, and the total length of the robot's trajectory.
Grippers that are easier to use should complete manipulation tasks with as little human guidance as necessary (e.g., fewer joystick inputs and shorter trajectory length).
We also administered a $7$-point Likert scale survey to assess the participants' subjective outcomes. 
Our questions are listed in Table~\ref{table:collab5}, and ask about how helpful, easy to use, and intuitive the system was.

\p{Shared Autonomy Results}
In Figure~\ref{fig:user2}B and Figure~\ref{fig:user2}C we present the results of this user study. 
Paired t-tests revealed that the overall grasping success rates were not significantly different for either human control or shared autonomy ($t(11) = -0.62$, $p=.55$, also see Table~\ref{table:collab4}). 
The main difference between conditions was the amount of human effort.
When working with shared autonomy, participants were able to complete the manipulation tasks with fewer joystick inputs ($t(11) = 13.65$, $p<.05$), and the shared autonomy system moved more efficiently to complete each grasp ($t(11) = 4.58$, $p<.05$).
In support of these objective results, users indicated a subjective preference for working with RISO under the shared autonomy framework (\fig{user2}C and Table~\ref{table:collab5}).
Post-hoc tests show that participants perceived the shared autonomy approach to be more helpful ($p<.05$), easier to use ($p<.05$), and overall preferable to the human controller ($p<.05$).
Viewed together, these objective and subjective results suggest that RISO grippers are not only capable of picking up a diverse range of objects, but also that we can reduce the human's effort when leveraging these RISO grippers.
\begin{figure*}[t]
	\centering
	\includegraphics[width=1\textwidth]{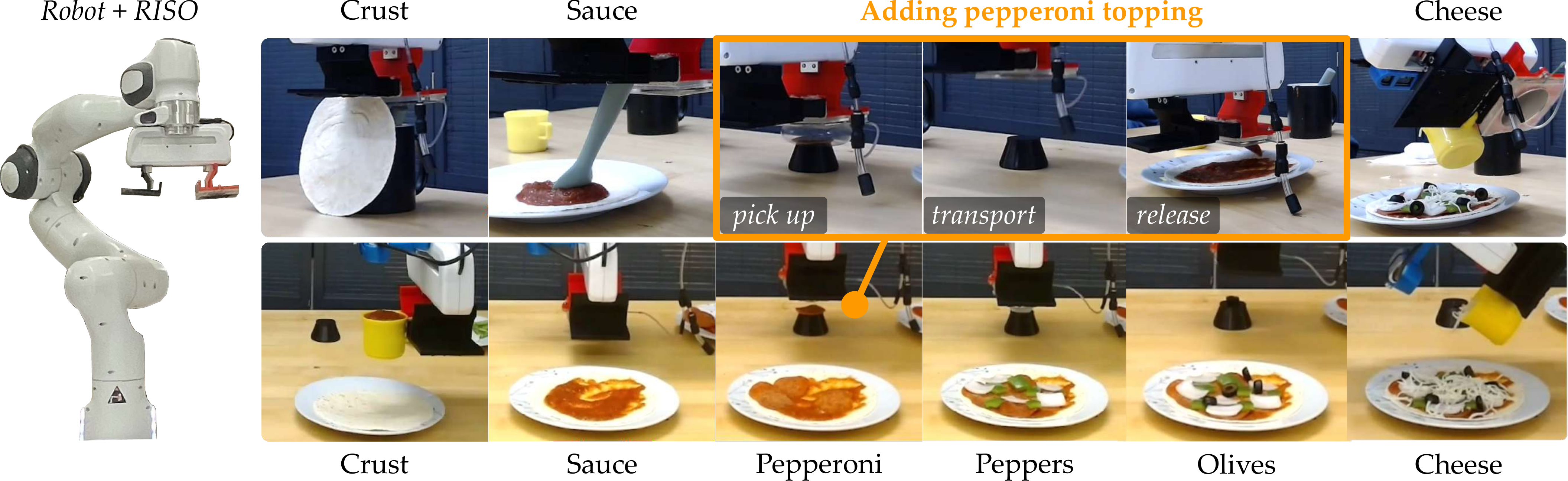}
	\caption{\textbf{RISO grasps and manipulates multiple food items to assemble a pizza.} A human teleoperated the robot arm and attached RISO gripper. Here we show snapshots of the resulting task from a side view (top row) and an overhead view (bottom row). The RISO used purely rigid grasps to manipulate the crust, spread the sauce, and pour the cheese. For small and numerous toppings the RISO used purely soft grasps: picking up, transporting, and releasing the pepperoni, peppers, and olives with the soft materials. See movie S10 for the overall task.}
	\label{fig:pizza}
\end{figure*}

\section{Discussion and Conclusion} \label{sec:discussion}

In this work we covered the surfaces of rigid end-effectors with a novel class of soft adhesives to form RISOs: RIgid-SOft grippers. 
Our underlying hypothesis was that RISOs would enhance grasping capabilities by both combining and decoupling rigid and soft mechanisms.
Because the rigid and soft components were \textit{combined}, RISO could perform rigid-soft manipulation, and because the components were \textit{decoupled}, RISO could also perform purely rigid or purely soft grasps.
We experimentally found that this combination enabled robots to overcome the fundamental limitations of each individual component.
In our user studies robot arms and human operators successfully utilized RISOs to pick up, move, and release a wide range of objects: from flat items to 3D objects, from single items to multiple objects, and from small items to large objects.
These enhanced grasping capabilities are practically useful in settings such as food processing, parts manufacturing, and assistive caregiving.
We demonstrate one practical application in Figure~\ref{fig:pizza}, where a human operator utilized our RISO gripper to assemble a pizza. 
This required grasping large, flat objects (the crust), manipulating tools (spreading the sauce and cheese), and picking up, transporting, and releasing small, irregular food items (the pepperoni, peppers, and olives).
Because RISO can perform grasps along a spectrum from purely rigid to purely soft, the robot completed each of these diverse manipulation tasks with a single RISO gripper.
Overall, we have demonstrated that RISOs provide exceptional gripping range (rapidly manipulating objects across a $1$ million times range in mass), and that robots and humans can effectively utilize RISO grippers within unstructured environments (with higher grasping success and user ratings as compared to state-of-the-art alternatives).

Another outcome of our studies was that the control scheme impacts the objective and subjective performance of the RISO gripper.
This demonstrates that the way the human and robot are integrated with the gripper can have an effect on gripper outcomes.
In particular, adhesion effectiveness is often characterized in controlled environments (i.e. static loading with precisely controlled alignment on testing machines) and considered to be a function of the adhesive alone \cite{croll2019switchable,jagota2011adhesion,creton2016fracture}.
However, when the adhesive or gripper is integrated with a robot arm and human user, this system now experiences variable inputs and motions that can significantly impact the functional adhesion capabilities.
From the soft material viewpoint, this signifies the importance of designing in tolerance to misalignment and applied forces, consideration of adhesion switching speed and the ability to maintain high forces during the full object manipulation process, and integration of soft adhesives into larger robotic systems.
\section*{Acknowledgments}

\p{Funding}
This work was supported in part by the National Science Foundation (Grant \#2205241).

\p{Data} Code associated with this study is available in the Supplementary Materials. All remaining data is available upon request.

\bibliographystyle{spmpsci}
\bibliography{citations}

\begin{thebibliography}{10}
\providecommand{\url}[1]{{#1}}
\providecommand{\urlprefix}{URL }
\expandafter\ifx\csname urlstyle\endcsname\relax
  \providecommand{\doi}[1]{DOI~\discretionary{}{}{}#1}\else
  \providecommand{\doi}{DOI~\discretionary{}{}{}\begingroup \urlstyle{rm}\Url}\fi

\bibitem{alizadehyazdi2020electrostatic}
Alizadehyazdi, V., Bonthron, M., Spenko, M.: An electrostatic/gecko-inspired adhesives soft robotic gripper.
\newblock IEEE Robotics and Automation Letters \textbf{5}(3), 4679--4686 (2020)

\bibitem{argall2018autonomy}
Argall, B.D.: Autonomy in rehabilitation robotics: {A}n intersection.
\newblock Annual Review of Control, Robotics, and Autonomous Systems \textbf{1}, 441--463 (2018)

\bibitem{autumn2002evidence}
Autumn, K., Sitti, M., Liang, Y.A., Peattie, A.M., Hansen, W.R., Sponberg, S., Kenny, T.W., Fearing, R., Israelachvili, J.N., Full, R.J.: Evidence for van der {W}aals adhesion in gecko setae.
\newblock Proceedings of the National Academy of Sciences \textbf{99}(19), 12252--12256 (2002)

\bibitem{barron2023unified}
Barron~III, E.J., Williams, E.T., Tutika, R., Lazarus, N., Bartlett, M.D.: A unified understanding of magnetorheological elastomers for rapid and extreme stiffness tuning.
\newblock RSC Applied Polymers  (2023)

\bibitem{bartlett2023peel}
Bartlett, M.D., Case, S.W., Kinloch, A.J., Dillard, D.A.: Peel tests for quantifying adhesion and toughness: {A} review.
\newblock Progress in Materials Science \textbf{137}, 101086 (2023)

\bibitem{bartlett2012looking}
Bartlett, M.D., Croll, A.B., King, D.R., Paret, B.M., Irschick, D.J., Crosby, A.J.: Looking beyond fibrillar features to scale gecko-like adhesion.
\newblock Advanced Materials \textbf{24}(8), 1078--1083 (2012)

\bibitem{birglen2018statistical}
Birglen, L., Schlicht, T.: A statistical review of industrial robotic grippers.
\newblock Robotics and Computer-Integrated Manufacturing \textbf{49}, 88--97 (2018)

\bibitem{brown2010universal}
Brown, E., Rodenberg, N., Amend, J., Mozeika, A., Steltz, E., Zakin, M.R., Lipson, H., Jaeger, H.M.: Universal robotic gripper based on the jamming of granular material.
\newblock Proceedings of the National Academy of Sciences \textbf{107}(44), 18809--18814 (2010)

\bibitem{cacucciolo2022peeling}
Cacucciolo, V., Shea, H., Carbone, G.: Peeling in electroadhesion soft grippers.
\newblock Extreme Mechanics Letters \textbf{50}, 101529 (2022)

\bibitem{chen2023soft}
Chen, H., Zhu, J., Cao, Y., Xia, Z., Chai, Z., Ding, H., Wu, Z.: Soft-rigid coupling grippers: {C}ollaboration strategies and integrated fabrication methods.
\newblock Science China Technological Sciences  (2023)

\bibitem{coulson2022versatile}
Coulson, R., Stabile, C.J., Turner, K.T., Majidi, C.: Versatile soft robot gripper enabled by stiffness and adhesion tuning via thermoplastic composite.
\newblock Soft Robotics \textbf{9}(2), 189--200 (2022)

\bibitem{creton2016fracture}
Creton, C., Ciccotti, M.: Fracture and adhesion of soft materials: a review.
\newblock Reports on Progress in Physics \textbf{79}(4), 046601 (2016)

\bibitem{croll2019switchable}
Croll, A.B., Hosseini, N., Bartlett, M.D.: Switchable adhesives for multifunctional interfaces.
\newblock Advanced Materials Technologies \textbf{4}(8) (2019)

\bibitem{DobotMagician}
{Dobot Magician}: {D}obot {M}agician {S}oft{G}ripper - 3 {F}ingers.
\newblock \url{https://www.soft-gripping.shop/en/dobot-magician-softgripper-3-fingers.html}

\bibitem{frankaFrankaEmika}
{Franka Robotics}: Franka {R}esearch 3.
\newblock \url{https://www.franka.de/}

\bibitem{gafer2020quad}
Gafer, A., Heymans, D., Prattichizzo, D., Salvietti, G.: The quad-spatula gripper: {A} novel soft-rigid gripper for food handling.
\newblock In: IEEE International Conference on Soft Robotics, pp. 39--45 (2020)

\bibitem{glick2018soft}
Glick, P., Suresh, S.A., Ruffatto, D., Cutkosky, M., Tolley, M.T., Parness, A.: A soft robotic gripper with gecko-inspired adhesive.
\newblock IEEE Robotics and Automation Letters \textbf{3}(2), 903--910 (2018)

\bibitem{guo2020self}
Guo, X.Y., Li, W.B., Gao, Q.H., Yan, H., Fei, Y.Q., Zhang, W.M.: Self-locking mechanism for variable stiffness rigid--soft gripper.
\newblock Smart Materials and Structures \textbf{29}(3) (2020)

\bibitem{hao2020multimodal}
Hao, Y., Biswas, S., Hawkes, E.W., Wang, T., Zhu, M., Wen, L., Visell, Y.: A multimodal, enveloping soft gripper: {S}hape conformation, bioinspired adhesion, and expansion-driven suction.
\newblock IEEE Transactions on Robotics pp. 350--362 (2020)

\bibitem{hernandez2023current}
Hernandez, J., Sunny, M.S.H., Sanjuan, J., Rulik, I., Zarif, M.I.I., Ahamed, S.I., Ahmed, H.U., Rahman, M.H.: Current designs of robotic arm grippers: {A} comprehensive systematic review.
\newblock Robotics \textbf{12}(1), 5 (2023)

\bibitem{hu2021soft}
Hu, Q., Dong, E., Sun, D.: Soft gripper design based on the integration of flat dry adhesive, soft actuator, and microspine.
\newblock IEEE Transactions on Robotics \textbf{37}(4), 1065--1080 (2021)

\bibitem{jagota2011adhesion}
Jagota, A., Hui, C.Y.: Adhesion, friction, and compliance of bio-mimetic and bio-inspired structured interfaces.
\newblock Materials Science and Engineering: R: Reports \textbf{72}(12), 253--292 (2011)

\bibitem{jain2019probabilistic}
Jain, S., Argall, B.: Probabilistic human intent recognition for shared autonomy in assistive robotics.
\newblock ACM Transactions on Human-Robot Interaction \textbf{9}(1), 1--23 (2019)

\bibitem{javdani2018shared}
Javdani, S., Admoni, H., Pellegrinelli, S., Srinivasa, S.S., Bagnell, J.A.: Shared autonomy via hindsight optimization for teleoperation and teaming.
\newblock The International Journal of Robotics Research \textbf{37}(7), 717--742 (2018)

\bibitem{jeon2020reward}
Jeon, H.J., Milli, S., Dragan, A.: Reward-rational (implicit) choice: {A} unifying formalism for reward learning.
\newblock Advances in Neural Information Processing Systems pp. 4415--4426 (2020)

\bibitem{jiang2022review}
Jiang, P., Ergu, D., Liu, F., Cai, Y., Ma, B.: A review of {Y}olo algorithm developments.
\newblock Procedia Computer Science \textbf{199}, 1066--1073 (2022)

\bibitem{jonnavittula2022sari}
Jonnavittula, A., Mehta, S.A., Losey, D.P.: {SARI: S}hared autonomy across repeated interaction.
\newblock ACM Transactions on Human-Robot Interaction  (2024)

\bibitem{keely2024kiri}
Keely, M.N., Nemlekar, H., Losey, D.P.: {Kiri-Spoon: A} soft shape-changing utensil for robot-assisted feeding.
\newblock arXiv preprint arXiv:2403.05784  (2024)

\bibitem{lee2023bioinspired}
Lee, C., Shi, H., Jung, J., Zheng, B., Wang, K., Tutika, R., Long, R., Lee, B.P., Gu, G.X., Bartlett, M.D.: Bioinspired materials for underwater adhesion with pathways to switchability.
\newblock Cell Reports Physical Science \textbf{4}(10) (2023)

\bibitem{li2022stiffness}
Li, L., Xie, F., Wang, T., Wang, G., Tian, Y., Jin, T., Zhang, Q.: Stiffness-tunable soft gripper with soft-rigid hybrid actuation for versatile manipulations.
\newblock Soft Robotics \textbf{9}(6), 1108--1119 (2022)

\bibitem{losey2022learning}
Losey, D.P., Jeon, H.J., Li, M., Srinivasan, K., Mandlekar, A., Garg, A., Bohg, J., Sadigh, D.: Learning latent actions to control assistive robots.
\newblock Autonomous Robots \textbf{46}(1), 115--147 (2022)

\bibitem{mehta2023riso}
Mehta, S.A., Kim, Y., Hoegerman, J., Bartlett, M.D., Losey, D.P.: Riso: {C}ombining rigid grippers with soft switchable adhesives.
\newblock In: IEEE International Conference on Soft Robotics (2023)

\bibitem{nasab2017soft}
Nasab, A.M., Sabzehzar, A., Tatari, M., Majidi, C., Shan, W.: A soft gripper with rigidity tunable elastomer strips as ligaments.
\newblock Soft robotics \textbf{4}(4), 411--420 (2017)

\bibitem{park2018hybrid}
Park, W., Seo, S., Bae, J.: A hybrid gripper with soft material and rigid structures.
\newblock IEEE Robotics and Automation Letters \textbf{4}(1), 65--72 (2018)

\bibitem{peng2024enhanced}
Peng, Z., Liu, D., Song, X., Wang, M., Rao, Y., Guo, Y., Peng, J.: The enhanced adaptive grasping of a soft robotic gripper using rigid supports.
\newblock Applied System Innovation \textbf{7}(1), 15 (2024)

\bibitem{piazza2019century}
Piazza, C., Grioli, G., Catalano, M., Bicchi: A century of robotic hands.
\newblock Annual Review of Control, Robotics, and Autonomous Systems \textbf{2}, 1--32 (2019)

\bibitem{robotiqProductsGrippers}
Robotiq: {P}roducts: {G}rippers, {C}amera and {F}orce {T}orque {S}ensors - {R}obotiq.
\newblock \url{https://robotiq.com/products}

\bibitem{rojas2016gr2}
Rojas, N., Ma, R.R., Dollar, A.M.: The {GR2 gripper: A}n underactuated hand for open-loop in-hand planar manipulation.
\newblock IEEE Transactions on Robotics \textbf{32}(3), 763--770 (2016)

\bibitem{ruotolo2021grasping}
Ruotolo, W., Brouwer, D., Cutkosky, M.R.: From grasping to manipulation with gecko-inspired adhesives on a multifinger gripper.
\newblock Science Robotics \textbf{6}(61), eabi9773 (2021)

\bibitem{sanneman2021state}
Sanneman, L., Fourie, C., Shah, J.A.: The state of industrial robotics: {E}merging technologies, challenges, and key research directions.
\newblock Foundations and Trends in Robotics \textbf{8}(3), 225--306 (2021)

\bibitem{shintake2018soft}
Shintake, J., Cacucciolo, V., Floreano, D., Shea, H.: Soft robotic grippers.
\newblock Advanced Materials \textbf{30}(29) (2018)

\bibitem{shintake2016versatile}
Shintake, J., Rosset, S., Schubert, B., Floreano, D., Shea, H.: Versatile soft grippers with intrinsic electroadhesion based on multifunctional polymer actuators.
\newblock Advanced Materials \textbf{28}(2), 231--238 (2016)

\bibitem{song2017controllable}
Song, S., Drotlef, D.M., Majidi, C., Sitti, M.: Controllable load sharing for soft adhesive interfaces on three-dimensional surfaces.
\newblock Proceedings of the National Academy of Sciences \textbf{114}(22), E4344--E4353 (2017)

\bibitem{song2014geckogripper}
Song, S., Majidi, C., Sitti, M.: Geckogripper: {A} soft, inflatable robotic gripper using gecko-inspired elastomer micro-fiber adhesives.
\newblock In: IEEE/RSJ International Conference on Intelligent Robots and Systems, pp. 4624--4629 (2014)

\bibitem{song2014soft}
Song, S., Sitti, M.: Soft grippers using micro-fibrillar adhesives for transfer printing.
\newblock Advanced Materials \textbf{26}(28), 4901--4906 (2014)

\bibitem{swift2020active}
Swift, M.D., Haverkamp, C.B., Stabile, C.J., Hwang, D., Plaut, R.H., Turner, K.T., Dillard, D.A., Bartlett, M.D.: Active membranes on rigidity tunable foundations for programmable, rapidly switchable adhesion.
\newblock Advanced Materials Technologies \textbf{5}(11), 2000676 (2020)

\bibitem{tatari2018dynamically}
Tatari, M., Mohammadi~Nasab, A., Turner, K.T., Shan, W.: Dynamically tunable dry adhesion via subsurface stiffness modulation.
\newblock Advanced Materials Interfaces \textbf{5}(18), 1800321 (2018)

\bibitem{tian2019gecko}
Tian, H., Li, X., Shao, J., Wang, C., Wang, Y., Tian, Y., Liu, H.: Gecko-effect inspired soft gripper with high and switchable adhesion for rough surfaces.
\newblock Advanced Materials Interfaces \textbf{6}(18), 1900875 (2019)

\bibitem{wan2023tunable}
Wan, G., Tang, Y., Turner, K.T., Zhang, T., Shan, W.: Tunable dry adhesion of soft hollow pillars through sidewall buckling under low pressure.
\newblock Advanced Functional Materials \textbf{33}(2), 2209905 (2023)

\bibitem{webb2024wearable}
Webb, H., Chanrungmaneekul, P., Yuan, S., Hang, K.: Wearable roller rings to enable robot dexterous in-hand manipulation through active surfaces.
\newblock arXiv preprint arXiv:2403.13132  (2024)

\bibitem{wei2016novel}
Wei, Y., Chen, Y., Ren, T., Chen, Q., Yan, C., Yang, Y., Li, Y.: A novel, variable stiffness robotic gripper based on integrated soft actuating and particle jamming.
\newblock Soft Robotics \textbf{3}(3), 134--143 (2016)

\bibitem{zhang2020state}
Zhang, B., Xie, Y., Zhou, J., Wang, K., Zhang, Z.: State-of-the-art robotic grippers, grasping and control strategies, as well as their applications in agricultural robots: {A} review.
\newblock Computers and Electronics in Agriculture \textbf{177}, 105694 (2020)

\end{thebibliography}

\appendix

\section*{Supplementary Materials}

\begin{figure*}[t]
	\centering
	\includegraphics[width=\textwidth]{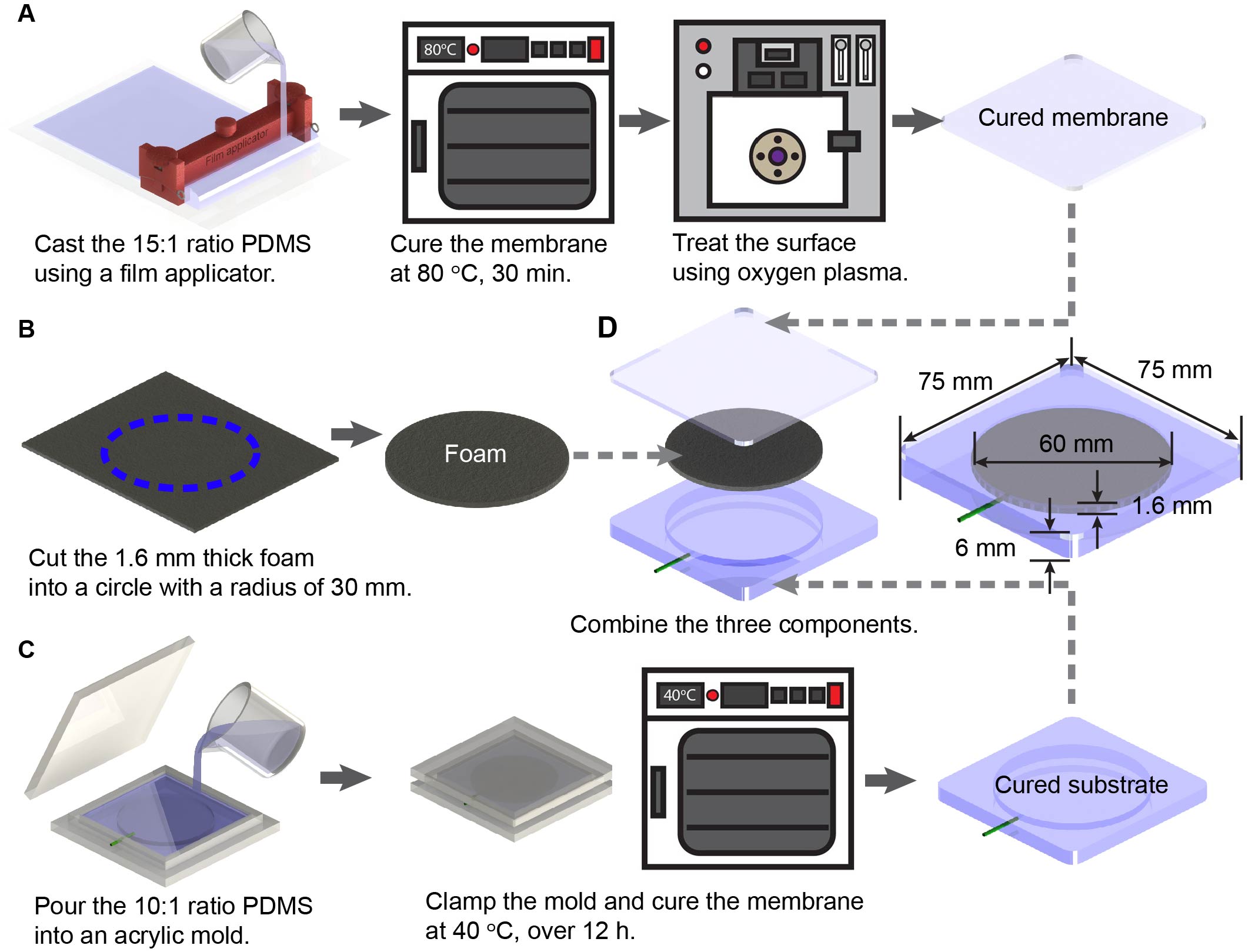}
	\caption{\textbf{Fabrication flow for the soft adhesive.} Fabrication process for the (\textbf{A}) membrane, (\textbf{B}) foam, (\textbf{C}) underlying substrate, and (\textbf{D}) integrated soft adhesive. The soft switchable adhesive has three subcomponents: a soft elastomeric Polydimethylsiloxane membrane (PDMS, Sylgard 184, Dow), 1.6 mm polyurethane foam foundation (Poron Very Soft 20 pcf Microcellular Polyurethane, Rogers Corporation), and an underlying PDMS substrate (PDMS, Sylgard 184, Dow). }
 \label{fig:SMAdhFab}
\end{figure*}

\begin{figure*}[t]
	\centering
	\includegraphics[width=\textwidth]{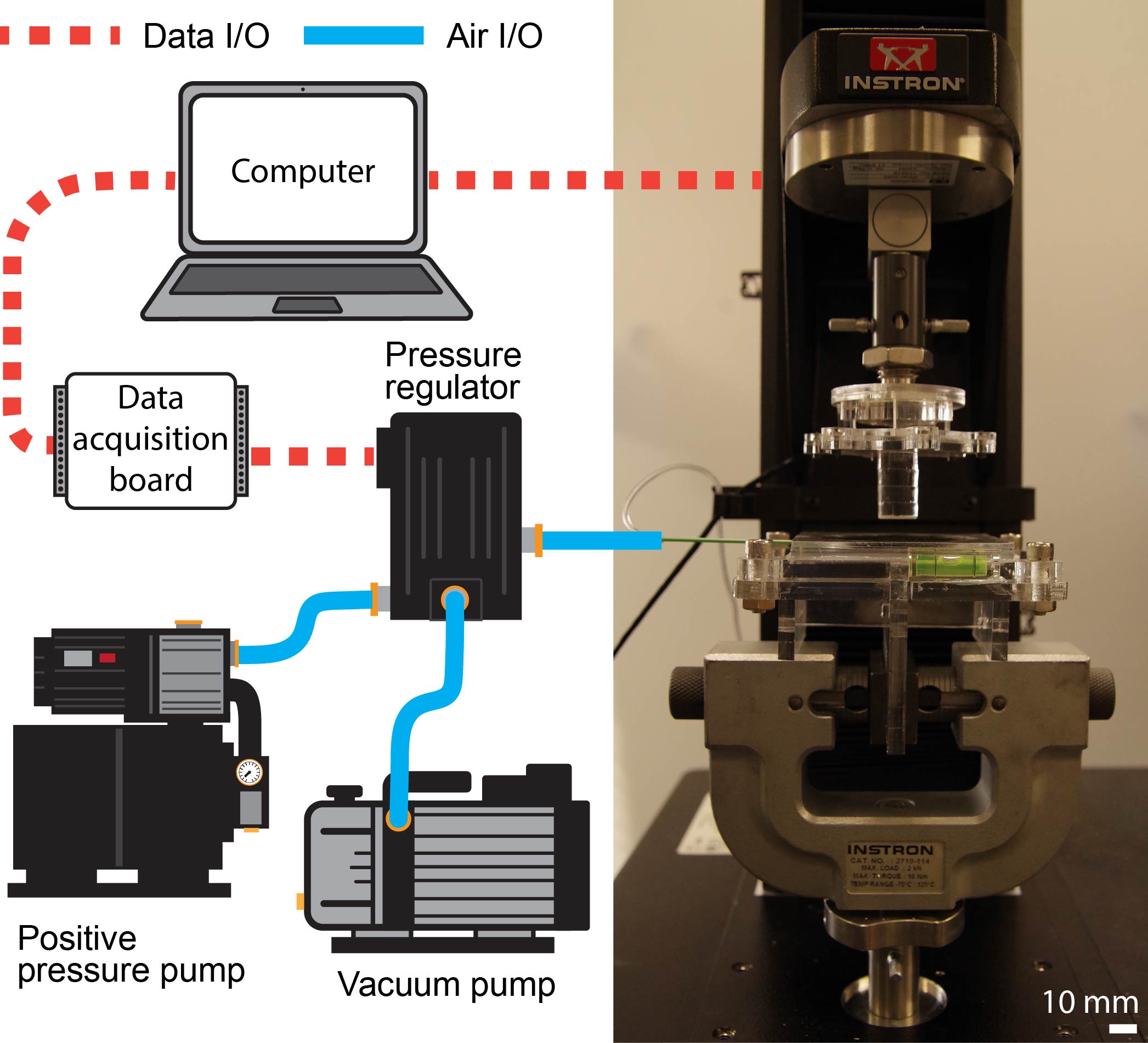}
	\caption{\textbf{Experimental setup for adhesion characterization.} The indentation experiment utilizes a pressure regulator to control the positive (1.5 kPa), neutral, or negative pneumatic pressure (-85 kPa) which is synchronized to the mechanical testing machine through a data acquisition board controlled by a computer. Tests are performed by displacing the sample until a preload of 25 kPa is reached, the preload is held for 5 seconds, and then the sample is retracted with a detachment speed of 10 mm/min. All substrates in Figures \ref{fig:indentertest} and \ref{fig:differentsurface} are acrylic and are manufactured by laser cutting. Indenters with different curvatures in Figure \ref{fig:differentsurface}A are cut from a hemispherical acrylic structure with a radius of 7.5 mm, and the indenters with different line distances are engraved with a power and speed of 30\% (PLS 6150, Universal Laser Systems). 
 }
 \label{fig:SMAdhSetup}
\end{figure*}

\begin{figure*}[t]
	\centering
	\includegraphics[width=\textwidth]{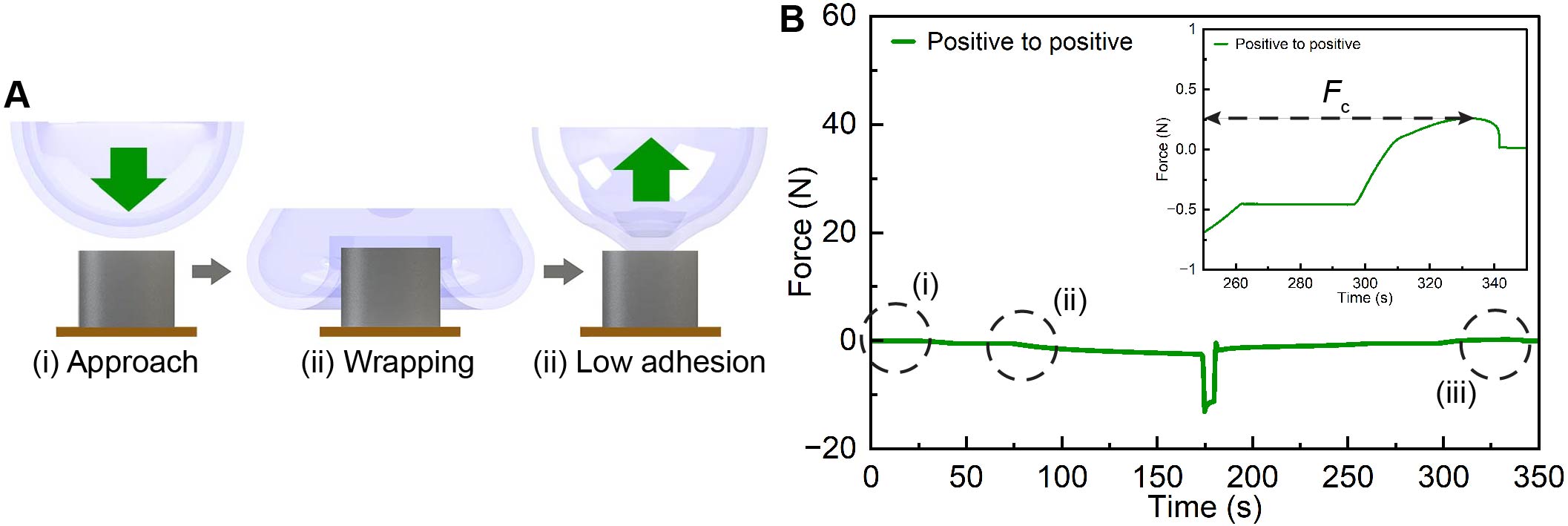}
	\caption{\textbf{Release mechanism of the soft adhesive.}
 (\textbf{A}) Schematics showing the approach, wrapping, and low adhesion state using an inflated membrane. (\textbf{B}) Force vs time plot from an indentation experiment where the inset shows the point where $F_c$ is measured.}
 \label{fig:SMAdhRelease}
\end{figure*}

\begin{figure*}[t]
	\centering
	\includegraphics[width=\textwidth]{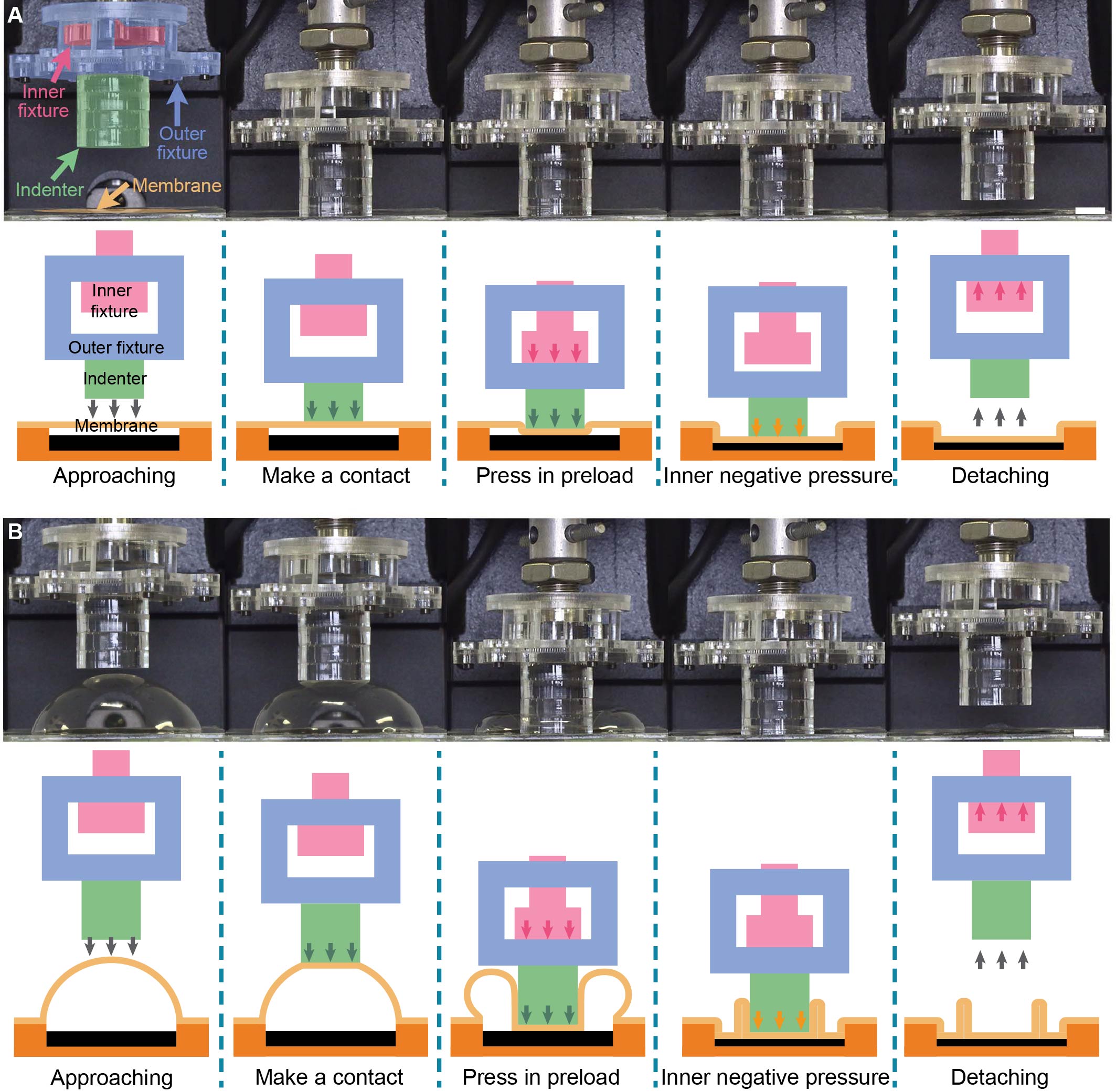}
	\caption{\textbf{Adhesion testing procedure.}
 (\textbf{A}) Images and schematics showing a sequence of the adhesion testing procedure of the neutral to negative condition and the (\textbf{B}) positive to negative condition. The scale bar is 10 mm.}
 \label{fig:SMAdhTest}
\end{figure*}

\begin{figure*}[t]
	\centering
	\includegraphics[width=\textwidth]{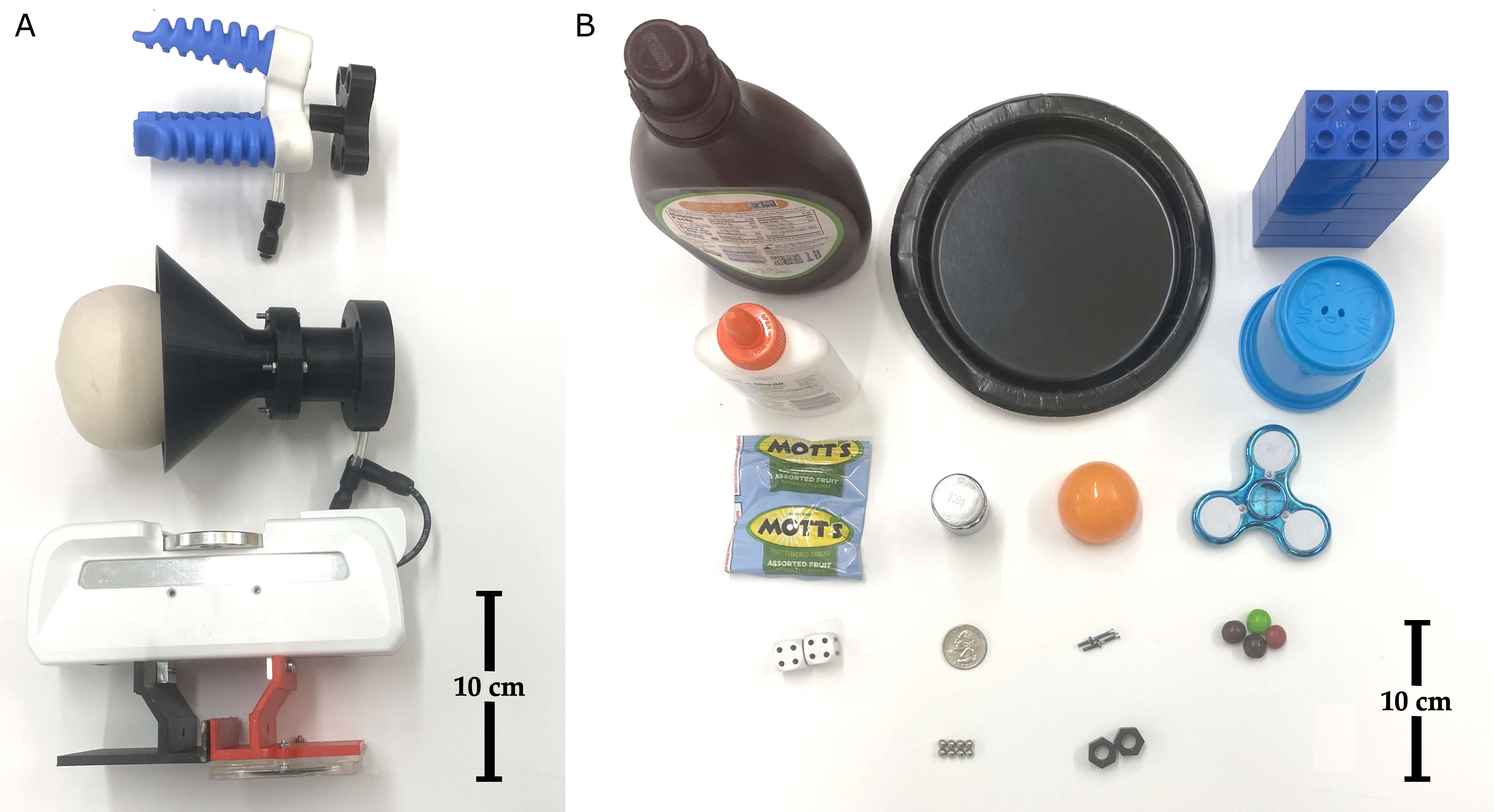}
	\caption{\textbf{Comparison of grippers and the dataset of objects.} \textbf{(A)} The SoftGripper \cite{DobotMagician}, the granular jamming gripper \cite{brown2010universal}, and our RISO Gripper. \textbf{(B)} The $15$ objects used in our experiments from Section~\ref{sec:r3} and Section~\ref{sec:r4}. For both images, a marker indicates the relative size of the items.}
     \label{fig:collab1}
\end{figure*}

\noindent \fig{SMAdhFab}. Fabrication flowchart for the soft adhesive. \smallskip\\
\fig{SMAdhSetup}. Experimental setup for adhesion characterization. \smallskip\\
\fig{SMAdhRelease}. Release mechanism of the soft adhesive. \smallskip\\
\fig{SMAdhTest}. Adhesion testing procedure. \smallskip\\
\fig{collab1}. RISO, SoftGripper, granular jamming gripper, and dataset of target objects. \smallskip\\
Table~\ref{table:collab3}. Questionnaire items and user responses when comparing RISO to other grippers. \smallskip\\
Table~\ref{table:collab2}. Grasping success rates with RISO and other state-of-the-art grippers. \smallskip\\
Table~\ref{table:collab5}. Questionnaire items and user responses when using RISOs with shared autonomy. \smallskip\\
Table~\ref{table:collab4}. RISO grasping success rates with human control and shared autonomy. \smallskip\\
Video. Summary video that visualizes the methods and experiments from the paper:\\  \url{https://youtu.be/du085R0gPFI} \smallskip\\
Code Repository. Link to an online repository with the code for a robot arm and attached RISO:\\
\url{https://github.com/VT-Collab/RISO_Gripper}

\clearpage

\clearpage

\clearpage

\clearpage

\clearpage

\clearpage
\begin{table*}[t]
\caption{\textbf{Questionnaire items and responses from our Likert scale survey comparing each type of gripper}. In Section~\ref{sec:r3} participants teleoperated the SoftGripper, the granular jamming gripper, and RISO to pickup a dataset of items. We used this survey to assess the user's subjective response to each different gripper. (Left) The questionnaire items listed below explore whether the robot picked up all the objects, whether it was easy to control the gripper, if the user could predict a successful grasp, and if they would prefer to use this gripper again in the future. (Right) After collecting all the responses, we first confirmed that the participants' answers to these questions were consistent using Cronbach's $\alpha$ (reliability $> 0.7$). We then used post-hoc analysis to see if the scores for RISO were higher than the scores for the SoftGripper or Granular gripper. Here a $p$-value of less than $0.05$ indicates that the participants scored RISO more highly (e.g., better at picking up objects), and that these differences in scores were statistically significant.}
\vspace{1em}
\label{tab:study1-likert}
\resizebox{\textwidth}{!}{%
\begin{tabular}{@{}llccc@{}}
\toprule
\multirow{2}{*}{\textbf{Questionnaire Item}} &
  \multirow{2}{*}{\textbf{Reliability}} &
  \multirow{2}{*}{\textbf{F(2, 22)}} &
  \multicolumn{2}{c}{\textbf{p-value}} \\ \cmidrule(l){4-5} 
                                                                           &  &  & SoftGripper & Granular \\ \midrule
The gripper picked up all the objects I wanted. &
  \multirow{2}{*}{0.859} &
  \multirow{2}{*}{33.754} &
  \multirow{2}{*}{p \textless 0.05*} &
  \multirow{2}{*}{p \textless 0.05*} \\
The gripper did not do a good job of picking up objects.                  &  &  &             &          \\ \midrule
I found it easy and intuitive to control the gripper. &
  \multirow{2}{*}{0.809} &
  \multirow{2}{*}{8.075} &
  \multirow{2}{*}{0.482} &
  \multirow{2}{*}{p \textless 0.05*} \\
I had trouble controlling the gripper to do what I wanted.                &  &  &             &          \\ \midrule
I could easily predict whether the gripper would pick up an object. &
  \multirow{2}{*}{0.896} &
  \multirow{2}{*}{8.052} &
  \multirow{2}{*}{p \textless 0.05*} &
  \multirow{2}{*}{p \textless 0.05*} \\
It was hard to guess if the gripper would pick up or drop an object. &  &  &             &          \\ \midrule
If I had to use this robot, I would like to use this gripper again. &
  \multirow{2}{*}{0.967} &
  \multirow{2}{*}{18.395} &
  \multirow{2}{*}{p \textless 0.05*} &
  \multirow{2}{*}{p \textless 0.05*} \\
I would not want to use this gripper in the future.                       &  &  &             &          \\ \bottomrule
\end{tabular}%
}
\label{table:collab3}
\end{table*}

\clearpage
\begin{table*}[t]
\caption{\textbf{Grasping success rate per object with RISO and other state-of-the-art grippers.} In a successful grasp the robot picks up, carries, and then releases the object(s). The success percentage is reported for three grippers: SoftGripper \cite{DobotMagician}, granular jamming gripper \cite{brown2010universal}, and our RISO (see Section~\ref{sec:r3}). In autonomous control the robot arm and gripper perform the task without a human-in-the-loop. By contrast, in human control a participant teleoperated the robot and gripper without autonomous assistance. We observe that the robot and human were able to successfully pick up and manipulate more objects with RISO as compared to the SoftGripper or granular jamming gripper.}
\vspace{1em}
\label{tab:study1}
\resizebox{\textwidth}{!}{%
\begin{tabular}{@{}lccc|ccc@{}}
\toprule
\multirow{3}{*}{\textbf{Objects}} &
  \multicolumn{3}{c|}{\textbf{Autonomous Control}} &
  \multicolumn{3}{c}{\textbf{Human Control}} \\ \cmidrule(l){2-7} 
 &
  \textbf{SoftGripper} &
  \textbf{Granular} &
  \textbf{RISO} &
  \textbf{SoftGripper} &
  \textbf{Granular} &
  \textbf{RISO} \\
 &
  \multicolumn{1}{l}{Success {[}\%{]}} &
  \multicolumn{1}{l}{Success {[}\%{]}} &
  \multicolumn{1}{l|}{Success {[}\%{]}} &
  \multicolumn{1}{l}{Success {[}\%{]}} &
  \multicolumn{1}{l}{Success {[}\%{]}} &
  \multicolumn{1}{l}{Success {[}\%{]}} \\ \midrule
\multicolumn{1}{l|}{Glue}           & 50.00\%  & 0.00\%   & 100.00\% & 25.00\%  & 41.67\%  & 100.00\% \\
\multicolumn{1}{l|}{Lego Tower}     & 90.00\%  & 100.00\% & 100.00\% & 91.67\%  & 83.33\%  & 100.00\% \\
\multicolumn{1}{l|}{Syrup Bottle}   & 100.00\% & 20.00\%  & 100.00\% & 100.00\% & 16.67\%  & 83.33\%  \\
\multicolumn{1}{l|}{Bearings}       & 3.75\%   & 0.00\%   & 97.50\%  & 9.79\%   & 0.00\%   & 89.58\%  \\
\multicolumn{1}{l|}{Cup}            & 100.00\% & 100.00\% & 100.00\% & 100.00\% & 100.00\% & 75.00\%  \\
\multicolumn{1}{l|}{Dice}           & 80.00\%  & 95.00\%  & 100.00\% & 100.00\% & 66.67\%  & 100.00\% \\
\multicolumn{1}{l|}{Fidget Spinner} & 90.00\%  & 100.00\% & 100.00\% & 66.67\%  & 91.67\%  & 100.00\% \\
\multicolumn{1}{l|}{Fruit Snacks}   & 90.00\%  & 0.00\%   & 100.00\% & 75.00\%  & 8.33\%   & 100.00\% \\
\multicolumn{1}{l|}{Hemisphere}     & 100.00\% & 100.00\% & 100.00\% & 100.00\% & 100.00\% & 100.00\% \\
\multicolumn{1}{l|}{Nuts}           & 75.00\%  & 95.00\%  & 90.00\%  & 95.83\%  & 54.17\%  & 79.17\%  \\
\multicolumn{1}{l|}{Paper Plate}    & 10.00\%  & 0.00\%   & 90.00\%  & 66.67\%  & 8.33\%   & 75.00\%  \\
\multicolumn{1}{l|}{Quarter}        & 20.00\%  & 30.00\%  & 100.00\% & 0.00\%   & 0.00\%   & 75.00\%  \\
\multicolumn{1}{l|}{Screws}         & 10.00\%  & 0.00\%   & 25.00\%  & 8.33\%   & 0.00\%   & 58.33\%  \\
\multicolumn{1}{l|}{Skittles}       & 30.00\%  & 22.50\%  & 95.00\%  & 31.25\%  & 8.33\%   & 97.92\%  \\
\multicolumn{1}{l|}{Weight}         & 0.00\%   & 100.00\% & 100.00\% & 0.00\%   & 33.33\%  & 91.67\%  \\ \bottomrule
\end{tabular}%
}
\label{table:collab2}
\end{table*}

\clearpage

\begin{table*}[t]
\caption{\textbf{Questionnaire items and responses from the Likert scale survey with shared autonomy}. In Section~\ref{sec:r4} participants teleoperated a robot and attached RISO to manipulate $15$ household items. They either controlled the robot directly (using human control) or with some assistance (shared autonomy). After completing the manipulation task with each control strategy the participants answered the following questions. (Left) The questionnaire items explored whether the robot helped users to complete the task, if it was intuitive to control the robot and the RISO gripper, if the robot recognized the user's intent, if it was easy to leverage the RISO, and if the participants preferred using that control approach. (Right) To analyze these results we first grouped the items into five scales and tested their reliability using Cronbach's $\alpha$. If the responses were reliable (i.e., if $\alpha > 0.7$) then we proceeded to use paired t-tests to compare the means. The $p$-values indicate if the users preferred shared autonomy over human control, and an $*$ denotes statistical significance. Participants perceived the RISO with shared autonomy as more helpful, better at recognizing their intent, easier to use, and preferable to the alternative. The participants responses to questions about intuitiveness were not consistent (i.e., not reliable enough to analyze).}
\label{tab:study2-likert}
\vspace{1em}
\resizebox{\textwidth}{!}{%
\begin{tabular}{@{}lccc@{}}
\toprule
\textbf{Questionnaire Item}                               & \textbf{Reliability} & \textbf{t(11)} & \textbf{p-value} \\ \midrule
The robot helped me to pick up objects.  & \multirow{2}{*}{0.943}  & \multirow{2}{*}{-3.115}   & \multirow{2}{*}{p \textless 0.05*} \\
The robot did not help me do the task.                   &                      &                &         \\ \midrule
I found it intuitive to control the robot arm and gripper.     & \multirow{2}{*}{-0.401} & \multirow{2}{*}{-}         & \multirow{2}{*}{-}                  \\
I was not sure how the robot would respond to my inputs. &                      &                &         \\ \midrule
The robot recognized what I was trying to do.                  & \multirow{2}{*}{0.904}  & \multirow{2}{*}{-3.872}   & \multirow{2}{*}{p \textless 0.05*} \\
The robot did not seem to learn my intent.               &                      &                &         \\ \midrule
This control mode made it easy to use the gripper.             & \multirow{2}{*}{0.878}  & \multirow{2}{*}{-3.83}    & \multirow{2}{*}{p \textless 0.05*} \\
It was hard to use the gripper the way I wanted.         &                      &                &         \\ \midrule
I preferred controlling the robot and gripper using this mode. & -                       & \multicolumn{1}{l}{-2.681} & p \textless 0.05*              \\ \bottomrule   
\end{tabular}%
}
\label{table:collab5}
\vspace{3em}
\end{table*}

\begin{table*}[t]
\caption{\textbf{Grasping success rate per object with human control and shared autonomy.} In Section~\ref{sec:r4} we compared how effectively participants were able to utilize RISO with human control and shared autonomy. Below we report the grasping success percentage per each object across all $12$ participants. Users were instructed to grasp the glue, Lego tower, and syrup bottle with RISO's rigid gripper, and participants used RISO's soft adhesive to grasp the remaining objects. We found that users had a similar success rate when working with human control and shared autonomy.}
\vspace{1em}
\label{tab:study2}
\centering
\resizebox{0.5\textwidth}{!}{%
\begin{tabular}{@{}lcc@{}}
\toprule
\multirow{2}{*}{\textbf{Objects}}   & \textbf{Human}   & \textbf{Shared}  \\
                                    & Success {[}\%{]} & Success {[}\%{]} \\ \midrule
\multicolumn{1}{l|}{Glue}           & 100.00\%         & 100.00\%         \\
\multicolumn{1}{l|}{Lego Tower}     & 100.00\%         & 100.00\%         \\
\multicolumn{1}{l|}{Syrup Bottle}   & 100.00\%         & 100.00\%         \\
\multicolumn{1}{l|}{Bearings}       & 91.67\%          & 93.75\%          \\
\multicolumn{1}{l|}{Cup}            & 75.00\%          & 58.33\%          \\
\multicolumn{1}{l|}{Dice}           & 91.67\%          & 100.00\%         \\
\multicolumn{1}{l|}{Fidget Spinner} & 75.00\%          & 83.33\%          \\
\multicolumn{1}{l|}{Fruit Snacks}   & 100.00\%         & 100.00\%         \\
\multicolumn{1}{l|}{Hemisphere}     & 100.00\%         & 100.00\%         \\
\multicolumn{1}{l|}{Nuts}           & 100.00\%         & 91.67\%          \\
\multicolumn{1}{l|}{Paper Plate}    & 75.00\%          & 75.00\%          \\
\multicolumn{1}{l|}{Quarter}        & 50.00\%          & 66.67\%          \\
\multicolumn{1}{l|}{Screws}         & 58.33\%          & 70.83\%          \\
\multicolumn{1}{l|}{Skittles}       & 87.50\%          & 87.50\%          \\
\multicolumn{1}{l|}{Weight}         & 100.00\%         & 100.00\%         \\ \bottomrule
\end{tabular}%
}
\label{table:collab4}
\end{table*}

\end{document}